\documentclass[preprint,12pt]{elsarticle}

\usepackage{amssymb}
\usepackage{amsmath}
\usepackage{array}
\usepackage{changepage}
\usepackage{booktabs}       
\usepackage{siunitx}        
\usepackage{threeparttable} 
\usepackage{multirow}       
\usepackage{xcolor}    
\usepackage{seqsplit}
\usepackage{array}
\usepackage[margin=1in]{geometry}
\usepackage{tcolorbox}
\usepackage{enumitem}
\usepackage{graphicx}
\usepackage{caption}
\usepackage{subcaption}
\usepackage{makecell}
\usepackage{tabularx}
\usepackage{pdflscape}
\usepackage{bbm}
\usepackage[T1]{fontenc}
\usepackage{url}

\journal{Ocean Engineering}

\begin{document}

\begin{frontmatter}

\title{Time-Aware Validation of Machine Learning Fuel Consumption Models: Evidence from 1\,Hz Operational Data, CCGS \textit{Sir Wilfrid Laurier}}

\author[1,2]{Samarasimha Reddy Chittamuru}
\author[1]{Ayhan Akinturk}
\author[1]{Allison Kennedy}
\author[1]{Joshua Barnes}
\author[2]{Matthew Hamilton}

\affiliation[1]{organization={National Research Council Canada, Ocean, Coastal and River Engineering},
            addressline={P.O. Box 12093}, 
            city={St. John's},
            postcode={A1B 3T5}, 
            state={NL},
            country={Canada}}

\affiliation[2]{organization={Department of Computer Science, Memorial University of Newfoundland},
            addressline={40 Arctic Avenue}, 
            city={St. John's},
            postcode={A1B 3X5}, 
            state={NL},
            country={Canada}}

\begin{abstract}
Ship fuel consumption (SFC) prediction supports vessel operation optimisation, emissions estimation, and decision support systems (DSS) for sustainable maritime transportation. Numerous data-driven fuel models have been developed over the past two decades, but a critical and often overlooked limitation lies in their validation practices: most studies evaluate performance using random train--test splits, which, applied to high-frequency records, admit temporal leakage and yield optimistic results that do not reflect deployment conditions. This paper examines that gap using time-aware evaluation, specifically Time Series Cross-Validation (TSCV) and Blocked TSCV (BTSCV). Using the Canadian Coast Guard Ship (CCGS) \textit{Sir Wilfrid Laurier} as a case study, six regression models and a physics baseline are tuned under three time-aware schemes and three feature configurations, then evaluated on a common chronological hold-out set drawn from
approximately 3.88 million steady-state 1\,Hz records.

Random Forest and XGBoost attain test $R^2$ of 0.99 and 0.98 under random partitioning, but $-0.36$ and $-0.10$ under chronological partitioning of the same records, whereas the penalised linear models improve. Ridge attains lower mean squared error than both ensembles in every time-aware comparison, each significant under a Wilcoxon signed-rank test with Benjamini--Hochberg correction. The physics baseline attains a hold-out RMSE of 86.09\,L\,h$^{-1}$, outperforming every tuned tree-based configuration under the same feature set while remaining below the linear models. Substituting speed through water with speed over ground, the only speed signal available before departure, reduces prediction accuracy within $\pm$15\% from 89--90\% to 83\%, quantifying the accuracy cost of a prediction that can be made ahead of the voyage. Rigorous time-aware validation is essential for trustworthy and generalisable fuel consumption models.
\end{abstract}


\end{frontmatter}

\section{Introduction}

Maritime shipping already moves over 80\% of world trade by volume. After a 2022 contraction, global maritime trade grew by 2.4\% in 2023 and followed by 2.2\% growth in 2024 (12.7 billion tonnes); for 2025–2029, United Nations Conference on Trade and Development (UNCTAD) projects average growth of 2.4\% in total seaborne trade and 2.7\% in containerized trade \cite{UNCTAD_RMT_2024,UNCTAD_RMT_2025}. From a fuel perspective, ships reporting to the IMO Data Collection System (DCS) used about 211~million tonnes of fuel in the 2023 reporting year, with 93.52\% being conventional oil-based fuels (HFO/LFO/Diesel–Gas oil) and the remainder comprising alternative fuels such as LNG (12.89~Mt), methanol (93{,}876~t), LPG, ethanol/ethane, and biofuels (390{,}846~t) \cite{IMO_DCS_2023}.  From 2016 to 2023, total tank-to-wake (TTW) greenhouse gas emissions increased by 12\% (compound annual growth rate 1.4\%), while shipping’s share of global anthropogenic CO\textsubscript{2}e\textsubscript{100} remained about 1.7\% (or 2.3\% of global anthropogenic CO\textsubscript{2}) \cite{ICCT_2025_ShippingShare}.

Since 1~January~2019, ships of 5{,}000 gross tonnage (GT) and above have been required under MARPOL Annex~VI (Regulation~22A) to collect and report annual fuel oil consumption data to the IMO Data Collection System (DCS) \cite{MEPC278_70,IMO_DCS_Page}. This mandate is generating large, growing corpora of operational ship fuel-consumption (SFC) records. With steady advances in machine learning and neural networks, this creates a clear opportunity to develop accurate, operationally useful FC models that support decision-support systems (DSS), voyage optimization, and emissions accounting. 

The availability of high-frequency operational data, however, places new demands on the protocols by which such models are evaluated. Because consecutive records are closely spaced in time and near-perfectly autocorrelated, random train--test partitioning assigns the temporal neighbours of nearly every test observation to the training set, so the resulting score quantifies interpolation within an already-observed operating regime rather than generalisation to conditions not encountered during training.

The implications of this behaviour for model evaluation are examined using high-frequency operational data acquired aboard the CCGS \textit{Sir Wilfrid Laurier}, a Canadian Coast Guard medium icebreaker, comprising approximately 3.88 million steady-state records sampled at 1\,Hz between August 2024 and June 2025 and paired with ERA5 hindcast environmental fields. Six regression models spanning penalised linear and tree-based families, together with a physics-based baseline, are tuned independently under three time-aware validation schemes and three feature configurations, then evaluated on a single common chronological hold-out set, so features, model classes, tuning protocol, and evaluation data remain fixed while the validation scheme alone is varied. Under random partitioning, the tree-based ensembles attain test $R^2$ of 0.99 and 0.98; under chronological partitioning of the same records, the corresponding values are $-0.36$ and $-0.10$, whereas the penalised linear models improve. Apparent model quality is therefore governed by the validation protocol at least as strongly as by the model class, and time-aware, leakage-resistant evaluation is advocated as the default for maritime fuel-consumption modelling.

\section{Literature Review}

\subsection{Evolution of Fuel Consumption Modeling}

Ship fuel consumption (SFC) modeling has evolved over the past two decades alongside advances in naval architecture, onboard data collection, and machine learning. Two recent reviews, by Fan et al. \cite{FAN2022112405} and Yan et al. \cite{yan2021data}, cover this domain from complementary angles. Fan et al. categorize SFC models into three types: white-box models (WBM), which rely entirely on physical mechanisms; black-box models (BBM), which are data-driven; and grey-box models (GBM), which combine physical laws with data-derived insights. Yan et al. frame fuel consumption modeling within maritime data analytics and operational decision-making, stressing model applicability, deployment, and real-world integration. Each paradigm trades off accuracy, interpretability, data availability, and computational cost. White-box and grey-box models remain important for applications with limited operational data, such as ship design, regulation, and energy efficiency assessment. This section reviews the development and application of such physically grounded models.

\subsubsection{Resistance Modeling in White-Box Approaches}

White-box models (WBM) of ship fuel consumption are grounded in the physics of naval architecture and marine hydrodynamics. They typically start by estimating the total hydrodynamic resistance on a vessel, which sets the basis for the propulsion power required under given operating conditions. Because they rely on measurable physical quantities rather than statistical inference, these models offer transparent, interpretable frameworks for assessing energy use and efficiency, and remain essential for ship design, performance simulation, and early-phase decision-making \cite{Tillig03102019}. This cubic-resistance formulation is adopted directly in Section 4.3 as a physics-based baseline, fitted to the vessel's own operational data and evaluated under the same validation protocols as the machine learning models. The following subsections cover the resistance components most commonly represented in white-box approaches, each contributing a distinct mechanism to a vessel's overall energy demand.

\paragraph{Calm-Water Resistance}
Calm-water resistance prediction is the foundation of most physically based ship performance models. The ITTC gave the first consistent treatment of viscous resistance with the ITTC 1957 model-ship correlation line \cite{Morrall1970_ITTC1957}, a practical method for estimating frictional drag on smooth hulls. The ITTC 1978 Performance Prediction Method \cite{ITTC_750203014_2017} followed, introducing form factors, correlation allowances, and propulsive coefficients to link model-scale experiments with full-scale performance. Holtrop and Mennen \cite{holtrop1982approximate} translated these principles into an empirical regression framework that let designers estimate power requirements directly from principal hull parameters. Kristensen and L{\"u}tzen \cite{kristensen2012prediction} later recalibrated the coefficients and added wind and sea-margin effects to fit contemporary vessel forms. These developments form the methodological baseline that most current white-box resistance and powering studies, including this one, build on.

\paragraph{Added Resistance in Waves}
In realistic sea conditions, vessels experience added resistance: extra drag from their motion response to incident waves. Researchers have studied this both theoretically and experimentally for decades. Salvesen \cite{salvesen1978added} was among the first to quantify it within linear seakeeping theory, showing how ship-wave interactions raise power demand in regular seas. Liu and Papanikolaou \cite{LIU2016211} later introduced a faster numerical framework based on potential-flow formulations, making added-resistance estimates practical across a wide range of hull geometries. More recently, the ITTC formalized these approaches in its recommended procedure for predicting power increase in irregular waves \cite{ITTC_750207022_2024}, now the primary reference for model testing and performance prediction under realistic sea states.

\paragraph{Effects of Wind and Ocean Currents}
Wind and ocean currents shape a vessel's effective resistance and power demand outside controlled test conditions. Aerodynamic drag on the exposed structure is routinely corrected in sea-trial evaluations following the ITTC Speed/Power Trials procedure \cite{ITTC_750401011_2024}, which provides reference coefficients for adjusting measured performance. Kristensen and L{\"u}tzen \cite{kristensen2012prediction} later folded these aerodynamic terms into their empirical power model, linking wind exposure to overall energy efficiency. Ocean currents are not a hydrodynamic resistance mechanism, but they change speed over ground and so alter the propulsion power needed to hold schedule or route efficiency. Cai et al. \cite{cai2014ship} showed that feeding real-time current data into voyage planning can yield fuel savings, especially on long routes.

\paragraph{Confinement Effects: Shallow and Restricted Waters}
Ship performance in shallow or confined waters differs from open seas because the surrounding flow field becomes highly constrained. Reduced under-keel clearance and limited channel cross-section accelerate the flow beneath and alongside the hull, raising pressure gradients and increasing wave-making. These changes add resistance, commonly described through blockage and squat effects. Schlichting's boundary-layer analyses \cite{Schlichting1979_BLT} explain this behaviour, showing how proximity to the seabed alters shear stress and pressure distribution around the hull. His work remains the theoretical basis for empirical corrections applied to model-scale predictions in shallow-water or port-approach conditions, where accurate power estimation depends strongly on depth and channel geometry.

\paragraph{Surface and Medium Condition: Fouling and Ice Resistance}
Hull condition and the surrounding medium change over time and can significantly alter hydrodynamic performance. Marine biofouling increases surface roughness and frictional resistance, with measurable penalties in fuel efficiency and emissions. Fonteinos et al. \cite{fonteinos2017ship} showed that shaft-torque and onboard sensor data can monitor this degradation, while Oliveira et al. \cite{oliveira2018effect} examined how hull geometry affects the magnitude of fouling-related drag. Song et al. \cite{song2020fouling} extended this analysis across different ship types and found fouling impacts vary by operation. In cold and ice-covered regions, additional resistance comes from mechanical interaction between the hull and sea ice. Huang et al. \cite{HUANG2021103057} developed an empirical model for this ice-induced resistance, supported by numerical simulations and full-scale data.

\subsubsection{Noon Report-Based Modeling}

White-box resistance models give physically grounded estimates of power demand, but they rely on idealized hull-form assumptions and controlled test conditions rather than real operational data. Noon reports were the first widely available data source to fill this gap. Compiled manually by ship crews, they provide daily summaries of operational, engine, and environmental parameters such as speed, draft, fuel consumption, and weather conditions. The data are coarse in temporal resolution and subject to observational inconsistencies, but they remain one of the most accessible and standardized sources of vessel performance information across global fleets. Their long time coverage and fleet-wide availability have made them a basis for empirical and data-driven fuel-consumption models, particularly when high-frequency sensor data are unavailable.

Beşikçi et al.~\cite{besikci2016ann} developed an Artificial Neural Network (ANN)–based decision support system to predict daily fuel consumption from noon report records collected under varying operational conditions. The model took ship speed, revolutions per minute (RPM), mean draft, trim, cargo onboard, and environmental descriptors such as wind and sea state as inputs. The ANN outperformed conventional multiple regression (MR) analysis, capturing nonlinear dependencies that often characterize sparse and heterogeneous maritime datasets. The resulting framework served as a practical onboard decision aid for voyage planning and fuel management.

Uyanık et al.~\cite{UYANIK2020102389} investigated fuel consumption prediction for a container vessel using noon report and engine logbook records. They compared Multiple Linear Regression, Ridge and LASSO Regression, Support Vector Regression, decision tree algorithms, and gradient-boosting methods on operational variables such as RPM, scavenge-air pressure, and shaft-torque indicators. The linear and Ridge models reached R² values approaching 0.99, reflecting the high internal consistency of the dataset and strong linear dependence among recorded parameters. This shows even simple algorithms can model well-structured noon report data, but it also exposes the limits of such low-frequency records in representing transient or high-dynamic operating conditions.

Zhou et al.~\cite{ZHOU2022255} proposed an adaptive hyperparameter-tuning framework to improve the reliability of fuel-consumption prediction models in complex maritime environments. Their approach used Bayesian optimization to calibrate model parameters while examining how environmental factors affect prediction accuracy. They evaluated four machine-learning models: Artificial Neural Networks (ANN), Support Vector Regression (SVR), Random Forest (RF), and Lasso Regression. The ANN gained the most from tuning, with R² increases of 0.08\% to 2.20\%. Adding environmental inputs improved the ANN and Lasso models but slightly hurt SVR and RF. The study shows that hyperparameter optimization combined with environmental context can produce more stable and interpretable noon-report-based models.

Gkerekos et al.~\cite{GKEREKOS2019106282} compared fuel-oil-consumption (FOC) models built with Support Vector Machines (SVM), Random Forest (RFR), Extra Trees (ETR), and Artificial Neural Networks (ANN). The study evaluated models trained on manually recorded noon reports alongside models trained on high-frequency measurements from Automated Data Logging and Monitoring (ADLM) systems. Ensemble methods, particularly ETR and RFR, delivered the highest accuracy in both data regimes. ADLM data improved model performance by roughly 7\% and shortened the observation period needed for reliable training, showing the trade-off between data granularity and availability.

Yan et al.~\cite{YAN2020101930} coupled Random Forest regression with speed-optimization analysis in a two-stage framework. First, a data-driven model trained on noon-report records predicted hourly fuel consumption. That model then fed an optimization routine that found the most fuel-efficient operating speed. Applied to real voyage data, it achieved fuel savings of 2\% to 7\%, showing a practical path from predictive modeling to direct energy-efficiency improvement.

Nguyen et al.~\cite{NGUYEN2023103261} built a testing framework to assess how robust ship fuel-consumption prediction (FCP) models are across multiple vessels. Using a 2.5-year dataset of noon reports from a global container fleet, they compared two multi-ship FCP architectures based on Extreme Gradient Boosting (XGB) and a Multi-Layer Perceptron (ANN). Rather than relying only on standard accuracy metrics, they examined how each model responded to variation in data quality and feature interdependence. No single algorithm dominated across all scenarios, which points to the value of cooperative or ensemble approaches. This work shifts attention from pure prediction accuracy toward robustness and interpretability under realistic operational uncertainty.

Chen et al.~\cite{CHEN2023114483} tested machine-learning models for estimating harbor-vessel fuel consumption by combining ship-specific operational variables with local meteorological inputs. They compared several algorithms, including Ridge Regression and Random Forest; Random Forest gave the most reliable results. In a case study of tugboat operations, adding weather-related parameters improved predictive accuracy by as much as 38.90\%, showing how much environmental context can affect model performance in port and harbor settings.

Noon-report-based modeling bridges physics-based and fully data-driven approaches: even coarse, manually logged data can support reliable fuel-consumption prediction. But its daily resolution limits the short-term operational detail it can capture, which motivates the shift toward higher-frequency sensor data examined next.

\subsubsection{Data-Driven Fuel Consumption Models Using High-Frequency Sensor Data}

Noon reports capture only a daily snapshot of vessel operation, too coarse to resolve the short-term dynamics that govern fuel use. Advances in onboard instrumentation and automated data-logging have closed this resolution gap: modern vessels now collect operational measurements at intervals of one second or less. These continuous data streams support performance assessment, maintenance scheduling, and increasingly, real-time operational decision-making. Researchers have built on this growing digital infrastructure to develop data-driven models that estimate fuel consumption with improved accuracy and temporal fidelity. Unlike noon reports, sensor-based datasets capture the short-term dynamics of propulsion, maneuvering, and environmental interaction that strongly shape a ship's energy profile.

The following section reviews representative studies that use high-frequency sensor data for fuel-consumption modeling. Each is examined by vessel type, data source and resolution, modeling technique, validation approach, and principal findings. Table~\ref{tab:highfreq_models} summarizes these studies and shows how validation design governs predictive reliability and determines whether a data-driven approach holds up in real-world operations.

\begin{table}[htbp]
\begin{adjustwidth}{-2.3cm}{}
\scriptsize
\centering
\caption{Overview of high-frequency sensor-based fuel consumption modeling studies.}
\label{tab:highfreq_models}
\begin{tabular}{
    p{2.6cm}
    p{2.1cm}
    p{3.2cm}
    p{2.5cm}
    p{3.4cm}
    p{3.6cm}
}
\toprule
\textbf{Literature} & \textbf{Ship Type} & \textbf{Data Source (Size)} & \textbf{Method(s)} & \textbf{Validation} & \textbf{Observations} \\
\midrule
Coraddu et al. (2017) & Handymax chemical/product tanker & Onboard 15\,s data aggregated to 15-min intervals; $\sim$100{,}000 data points & RLS, Lasso, RF & \textit{Random --} 30$\times$ random train--test splits across varying data sizes; Bayesian optimization for hyperparameter tuning & Compared WBM, BBM, and GBM for trim optimization. Used 30$\times$ random splits with Bayesian optimization for tuning. \\
\addlinespace[0.5em]
Bui-Duy et al. (2020) & Container ship & Operational + secondary (wind, TEU); 10{,}853 records over 2 years & Deep learning neural network (DLNN) & \textit{Random --} 80:20 train--test split; 7:1 internal validation & Model used in ATSP framework for route optimization; MAPE = 5.89\% \\
\addlinespace[0.5em]
Kim et al. (2021) & Container ship (13{,}000 TEU) & 1-min AMS data (6 months); size not specified & ANN, MLR, SVM, Ensemble & \textit{Random + voyage-level --} 70:30 random split + voyage-level evaluation & Evaluated both randomly and voyage-wise; emphasis on real-world deployment performance \\
\addlinespace[0.5em]
Agand et al. (2022) & Passenger ferry & 1-min sensor data (2019--2021); $\sim$1M records & MLR, DT, RF, ANN, XGBoost & \textit{Random --} 70:30 random split + 10-fold CV & XGBoost performed best; consistent route; 34 features used \\
\addlinespace[0.5em]
Xie et al. (2023) & Oil tanker (trial) & 1 Hz sensor data; 378{,}468 records (4.38 days), 496 features & XGBoost, RF, BP-ANN, MLR, polynomial regression & \textit{Random --} 10$\times$ random 70:30 splits, 10-fold CV & Applied Kwon method for acceleration/deceleration filtering; $R^2$ reached 0.9977 (XGBoost); robust to sensor noise and data quality issues \\
\addlinespace[0.5em]
Hu et al. (2023) & Bulk carrier & Flowmeter, navigation, meteorological sensors (300\,s); 24{,}300 points & Stacking (hybrid ML), trim optimization & \textit{Random --} random 80:20 split (5 seeds) & Two-voyage real-world case study; hybrid stacking model outperformed all single models; trim optimization reduced fuel/emissions by up to 1.82\% \\
\addlinespace[0.5em]
Papandreou et al. (2023) & VLCC (very large crude carrier) & Operational onboard sensors (speed, draft, engine data, weather, laden/ballast status; 15-min interval); 107{,}389 records & MPR, ANN, XGBoost & \textit{Random (stratified) --} 80:10:10 random split with seed; stratified; same data used for all models & Real deployment case, strong preprocessing and stratified splits across ML pipelines \\
\addlinespace[0.5em]
Lang et al. (2023a) & Chemical tanker & 1 Hz, downsampled to 15 min, 5 years of data; 63{,}093 records & XGBoost, ANN, SVR, Linear, GAM, polynomial regression & \textit{Random + voyage-level --} 3 unseen voyages (temporal); random train--test split & $R^2$/MAE not reported for validation voyages; learning curve tested for data-size impact \\
\addlinespace[0.5em]
Lang et al. (2023b) & PCTC & 1 Hz, downsampled to 10 min, 3 months of data; 8{,}632 records & Same as above & \textit{Random --} random train--test split only & Not enough data to use full voyages for validation \\
\addlinespace[0.5em]
Zhou et al. (2023) & Tuna seiners (Ship E and J) & $\sim$1 Hz averaged data; Ship E: Feb--May 2022 ($\sim$47{,}000 records), Ship J: Jun 2021--Jan 2022 ($\sim$42{,}000 records) & MLR, RF, Lasso, Bayesian, ANN & \textit{Time-aware --} 10$\times$ repeated 5-fold CV for training; 5$\times$2cv F-test for model comparison; testing on 3 months of unseen data & Accuracy $>$94\% (E), $>$92\% (J); $>$95\% and $>$90\% of samples within 15\% error, respectively \\
Fan et al. (2024) & Hybrid-power inland bulk carrier & Onboard sensor data; 5,897 raw samples, 3,376 after preprocessing; 16 features & DT, RF, XGBoost, SVM, ANN, GTB & \textit{Random --} 75:25 train--test split; random five-partition sample-size analysis & RF and XGBoost were best-performing; preprocessing and thermotechnical parameters had major impact on model accuracy \\
\bottomrule
\end{tabular}
\end{adjustwidth}
\end{table}

Agand et al.~\cite{AGAND2023115271} applied XGBoost, Random Forest, and MLP to predict fuel consumption for a passenger ferry using 1-minute resolution sensor data collected between 2019 and 2021. With around 1 million records and 34 features, they used a 70:30 train-test split plus 10-fold cross-validation; XGBoost performed best.

Coraddu et al.~\cite{CORADDU2017351} modeled fuel consumption for a Handymax chemical tanker using 15-second onboard sensor data aggregated to 15-minute intervals, applying Recursive Least Squares (RLS), Lasso regression, and Random Forest. They used 30 random train-test splits across varying data sizes and Bayesian optimization for hyperparameter tuning, and compared white-box and grey-box models focused on trim optimization.

Xie et al.~\cite{jmse11040738} used 1 Hz trial data over 4.38 days (378,468 records) from an oil tanker to evaluate XGBoost, RF, BP-ANN, and polynomial regression, filtering out acceleration/deceleration periods with the Kwon method. XGBoost performed best, reaching R² = 0.9977. They validated the models with 10 random 70:30 splits and 10-fold cross-validation.

Kim et al.~\cite{jmse9020137} predicted fuel consumption for a 13,000 TEU container ship using 1-minute AMS data collected over six months, implementing ANN, MLR, SVM, and ensemble models. They ran both random splitting and voyage-level validation to check deployment robustness, and emphasized real-world applicability and interpretability.

Hu et al.~\cite{HU2022110904} proposed a hybrid stacked ensemble model combined with trim optimization for a bulk carrier, using 5-minute fuel flowmeter and environmental data from two real-world voyages (about 24,300 data points), with five random 80:20 splits. The hybrid model outperformed individual learners and cut fuel consumption and emissions by up to 1.82\%.

Bui-Duy et al. (2020)~\cite{BUIDUY20211} built a deep learning model for route optimization in liner shipping, using 10,853 records over two years that included wind speed, direction, vessel capacity, and voyage characteristics. Trained with an 80:20 split and 7:1 internal validation, the model reached a MAPE of 5.89\% and was integrated into an asymmetric traveling salesman problem (ATSP) framework for optimal routing.

Papandreou et al.~\cite{PAPANDREOU2022110321} studied a VLCC using 15-minute onboard data with over 107,000 points, evaluating MPR, ANN, and XGBoost. They split the dataset into 80\% training, 10\% validation, and 10\% test with stratified random sampling, then used repeated 5-fold cross-validation and a 5$\times$2cv F-test to assess significance between models. XGBoost showed the strongest stability and accuracy.

Lang et al. (2023)~\cite{LANG2022110387} modeled ship propulsion power using full-scale operational data from two vessel types — a chemical tanker and a PCTC — comparing XGBoost, artificial neural networks (ANN), support vector regression (SVR), and statistical regression on predictive accuracy. Training and testing sets were randomly partitioned, but the authors also validated the models on three complete, unseen voyages from the chemical tanker dataset. XGBoost generalized best across these voyages, outperforming both the other ML models and a physics-based model, especially in high sea states. However, without explicit R² or MAE values for the validation set, it's hard to quantify how well generalization held up relative to training and testing performance.

Zhou et al.~\cite{ZHOU2023115509} built a two-step fuel consumption framework for fishing vessels: first predicting Speed Through Water (STW) from environmental and vessel data, then estimating Fuel Oil Consumption (FOC) using the predicted STW as an input. Using data from two tuna seiners (Ship E and Ship J), they applied a strict time-based split so the test data came from unseen future months (Feb–May 2022) — a realistic, deployment-ready test. They trained with 10-times repeated 5-fold cross-validation and compared model performance with a 5×2cv combined F-test. Rather than relying only on R², they introduced three accuracy measures on the unseen monthly test sets: Mean Absolute Error (MAE), Mean Accuracy, and the percentage of predictions within 15\% deviation from ground truth ("\% of Sample Accuracy $>$ 85\%"). Ship E reached over 94\% mean accuracy in every month, with over 95\% of predictions within 15\% error; Ship J reported 92.97 ± 0.69\% mean accuracy and 90\% of predictions within the same threshold. Testing on future, truly unseen data makes this one of the strongest evaluations of deployment reliability in the literature.

Fan et al.~\cite{FAN2024106946} developed machine learning models to predict fuel consumption for a hybrid-power inland bulk carrier , using 16 input features spanning navigational, environmental, and thermotechnical main-engine variables, with main-engine fuel consumption as the target. Of 5,897 collected samples, 3,376 remained after outlier screening and removal of null-value segments. They compared six methods — Decision Tree (DT), Random Forest (RF), Extreme Gradient Boosting (XGBoost), Support Vector Machine (SVM), Artificial Neural Network (ANN), and Gradient Tree Boosting (GTB)) — using a 7.5:2.5 train-test split, plus additional random five-partition analyses to study the effect of sample size. RF and XGBoost achieved the best performance, and adding thermotechnical engine parameters substantially improved accuracy.

As Table~\ref{tab:highfreq_models} shows, the dominant validation protocol in this literature is the random train--test split or standard $k$-fold cross-validation — appropriate for independent observations, but not designed to catch the temporal leakage that arises when highly autocorrelated, adjacent 1 Hz records get shuffled across train and test partitions. Kim et al., Lang et al., and Zhou et al. are the exceptions: they add voyage-level or genuinely time-aware testing alongside their primary random-split results, with Zhou et al. going furthest by testing on three months of chronologically later, unseen data. These three studies are the closest methodological precedents to the present work. Random splits and $k$-fold cross-validation were built for independent observations; applied to time-ordered records, they let adjacent samples inform one another across the split. This effect is modest for coarse daily data but grows as sampling frequency increases — which is why time-aware validation needs systematic treatment in the high-frequency setting examined here, not just a supplementary check.

\subsection{Research Gap and Objectives}

The studies reviewed in Section~2.1.3 show that high-frequency sensor data support accurate fuel-consumption prediction across a wide range of vessel types and model classes. They also reveal a consistent methodological pattern. As Table~\ref{tab:highfreq_models} shows, random train--test splitting and standard $k$-fold cross-validation remain the dominant evaluation protocols, and these are well-justified defaults: they are simple, widely understood, and appropriate for independent observations. Their statistical assumptions, however, are not satisfied by high-frequency operational records, in which consecutive observations are separated by a single second and are near-perfectly autocorrelated. Under random partitioning, a test sample is typically bordered on either side by training samples recorded moments earlier and later, so the reported test error reflects interpolation within an already-observed operating regime rather than genuine generalization to unseen conditions. This has direct consequences for model selection: a model favoured under such a protocol may not be the model best suited to deployment.

A small number of studies have moved beyond this default. Kim et al.~\cite{jmse9020137} and Lang et al.~\cite{LANG2022110387} supplemented random splits with voyage-level holdouts, and Zhou et al.~\cite{ZHOU2023115509} evaluated on three months of chronologically subsequent data. These contributions show the value of chronologically honest evaluation, but in each case the time-aware assessment functions as a supplementary check applied to a model already selected under random splitting. Whether the validation protocol itself determines which model is selected remains untested. Answering this requires holding features, model classes, hyperparameter tuning protocol, and test data fixed while varying only the validation scheme, then evaluating the resulting configurations on a common chronologically held-out period.

This study addresses that gap through the following objectives:

\begin{enumerate}
\item To evaluate three validation protocols—Sequential K-Fold, Custom Time  Series Cross-Validation, and Blocked Time Series Cross-Validation—by  independently tuning six regression models (MLR, Ridge, Lasso, ElasticNet,  Random Forest, and XGBoost) under each protocol and comparing the resulting  configurations on a single, common chronological hold-out set, quantifying  how much apparent model performance depends on the protocol used for selection.

\item To evaluate three feature configurations—speed over ground (SOG) alone,  speed through water (STW) alone, and their combination—quantifying the  predictive cost of restricting inputs to those available at the voyage-planning  stage, since STW is measured onboard and unavailable prior to departure while  SOG can be forecast in advance.

\item To fit a cubic speed–draft resistance formulation, consistent with  established white-box resistance relations, alongside the six machine  learning models under identical partitioning and metrics, providing a  physically grounded reference for the data-driven models.

\item To test performance differences for statistical significance using a  Wilcoxon signed-rank test on block-mean squared-error differentials, with  Benjamini–Hochberg correction across the full batch of comparisons, so that  reported rankings rest on tested differences rather than point estimates alone.

\item To interpret model behaviour through SHAP values for the tree-based  ensembles and standardised coefficients for the linear model, placed on a  common scale for cross-family comparison and linked to the underlying  resistance relations.
\end{enumerate}

\section{Data Description and Sources}

\subsection{Vessel Description and Data Overview}
The dataset analyzed in this study was collected aboard the \textit{CCGS Sir Wilfrid Laurier}, a Martha Black-class medium icebreaker operated by the Canadian Coast Guard (CCG), built in 1986 and homeported in Victoria, British Columbia~\cite{ccgsWiki2023,jmse10040522}. In 2024, the vessel underwent a Vessel Life Extension (VLE) refit that replaced all four main diesel engines~\cite{seaspanVLE2024,marinelogVLE2024}. Following the refit, the propulsion configuration consists of three Wärtsilä 8L26 main engines and one Caterpillar 3508 auxiliary engine. The Wärtsilä 26 series is a medium-speed diesel platform certified for marine diesel oil and biofuel blends~\cite{wartsila26}.

This refit matters for the dataset because it fixes a single, consistent propulsion configuration for the study period. The analysis uses only operational data collected after the engine replacement, spanning August 2024 to June 2025, so all records reflect the same engine configuration. The dataset integrates onboard sensor streams with hindcast environmental parameters, synchronized by timestamp and by GPS-based geospatial matching. The composition, preprocessing, and integration of these sources are detailed in the following subsections.

\subsubsection{Onboard Sensor Data}
Onboard measurements were acquired continuously during vessel operation via the OpDAQ\texttrademark\ monitoring system, which interfaces with the ship's flowmeter network to record engine-specific fuel consumption in real time. This system provided time-stamped fuel flow rates for the three \textit{Wärtsilä 8L26} main engines (ME1, ME2, ME3) and the auxiliary \textit{Caterpillar 3508} generator, forming the primary response variable for this study.

Navigational and environmental parameters were obtained from the vessel's instrumentation channels under the National Marine Electronics Association (NMEA) 0183 communication standard~\cite{nmea0183}, comprising vessel speed, heading, rudder status, and wind conditions. Relevant signals were extracted from the following NMEA sentences: Speed over Ground (VTG), Speed Relative to Water (VHW), True Heading (HDT), Wind Speed and Direction (MWV), and the Rudder Control System (DSC1). These channels characterize the vessel's hydrodynamic response and environmental exposure, forming the principal predictor set for the regression models developed in Section~4.

Table~\ref{tab:sensor_channels} summarizes the onboard sensor channels retained for this study, together with their measurement units and acquisition sources.

\begin{table}[htbp]
\centering
\small
\setlength{\tabcolsep}{4pt}
\renewcommand{\arraystretch}{0.95}
\caption{Onboard sensor channels integrated for fuel consumption modeling.}
\label{tab:sensor_channels}
\begin{threeparttable}
\begin{tabular}{@{}l l p{5.2cm} l l@{}}
\toprule
\textbf{Category} & \textbf{Column Name} & \textbf{Description} & \textbf{Unit} & \textbf{Source} \\
\midrule
\multirow{4}{*}{Fuel Flow}
 & \texttt{ME1\_Flow}      & Fuel flow rate, Main Engine 1        & L/h & OpDAQ \\
 & \texttt{ME2\_Flow}      & Fuel flow rate, Main Engine 2        & L/h & OpDAQ \\
 & \texttt{ME3\_Flow}      & Fuel flow rate, Main Engine 3        & L/h & OpDAQ \\
 & \texttt{Aux\_Gen\_Flow} & Fuel flow rate, auxiliary generator  & L/h & OpDAQ \\
\midrule
\multirow{2}{*}{Speed}
 & \texttt{OpDAQ GPS SOG}                    & Speed over ground (SOG)      & kn & VTG \\
 & \texttt{VHW Speed Relative to Water Kn}   & Speed through water (STW)    & kn & VHW \\
\midrule
\multirow{2}{*}{Wind}
 & \texttt{MWV Wind Speed} & Apparent wind speed                        & kn  & MWV \\
 & \texttt{MWV Wind Angle} & Apparent wind direction rel.\ heading      & deg & MWV \\
\midrule
\multirow{2}{*}{Draft}
 & \texttt{Fore\_Draft} & Draft reading at ship's bow& m & DAQ \\
 & \texttt{Aft\_Draft}  & Draft reading at ship's stern& m & DAQ \\
\midrule
Heading & \texttt{HDT Heading} & Heading relative to true north & deg & HDT \\
\midrule
\multirow{2}{*}{Rudder}
 & \texttt{DSC1 Rudder Angle (Actual)} & Actual rudder angle            & deg & DSC1 \\
 & \texttt{DSC1 Rudder Status} & Operational status of rudder  & --  & DSC1 \\
\bottomrule
\end{tabular}
\end{threeparttable}
\end{table}

\subsubsection{Metocean Data}
To account for the influence of environmental forcing on vessel performance, hindcast wave and swell parameters were obtained from the ERA5 reanalysis archive (fifth-generation ECMWF atmospheric reanalysis)~\cite{hersbach2023era5}, produced by the European Centre for Medium-Range Weather Forecasts (ECMWF) and distributed through the Copernicus Climate Data Store (CDS) under the Copernicus Climate Change Service (C3S). Data were retrieved programmatically via the \texttt{cdsapi} Python client~\cite{cdsapi2024} for automated, reproducible spatial--temporal extraction aligned with the vessel's voyage track. The ERA5 ocean wave dataset has a spatial resolution of 0.5$^{\circ}$ and a temporal resolution of one hour, interpolated to match the sampling frequency of the shipborne sensor data.

Table~\ref{tab:era5_params} summarizes the selected ERA5 parameters, which describe the sea state affecting ship resistance and propulsion demand: significant and swell wave heights, mean and peak periods, and propagation directions for both total and swell wave components.

\begin{table}[htbp]
\centering
\caption{ERA5 metocean parameters used in this study}
\label{tab:era5_params}
\begin{threeparttable}
\begin{tabular}{l p{1.6cm} p{7cm} l}
\toprule
\textbf{Parameter} & \textbf{Variable Code} & \textbf{Description} & \textbf{Unit} \\
\midrule
Significant wave height & \texttt{swh} & Average height of the highest one-third of ocean waves & m \\
Mean wave direction & \texttt{mwd} & Mean direction from which the combined wind-sea and swell wave field is propagating & degrees \\
Mean wave period & \texttt{mwp} & Mean period of the wave spectrum & s \\
Peak wave period & \texttt{pp1d} & Period corresponding to the maximum of the wave spectrum & s \\
\midrule
Significant swell height & \texttt{shts} & Average height of the swell component of waves & m \\
Mean swell direction & \texttt{mdts} & Direction from which the swell component is propagating & degrees \\
Mean swell period & \texttt{mpts} & Mean period of the swell component & s \\
\bottomrule
\end{tabular}
\end{threeparttable}
\end{table}

\subsection{Data Cleaning and Preparation}
The raw dataset comprised approximately 26 million records collected between August 2024 and June 2025 at a 1~Hz sampling rate. A multi-stage processing pipeline retained only high-quality, operationally consistent measurements for model development and evaluation, comprising signal integration, quality assurance, and steady-state filtering.

\subsubsection{Data Integration and Storage}
Onboard and hindcast datasets were consolidated into a unified time-series structure to enable synchronized analysis. ERA5 environmental parameters were matched to onboard sensor measurements by timestamp and vessel position. The integrated dataset was managed in an in-house PostgreSQL database~\cite{singh2025dataplatform}, supporting efficient querying, subsetting, and export for the modeling workflows described in the following sections.

\subsubsection{Data Quality Assurance and Filtering}

Only measurements meeting the highest data-quality standard were retained. For NMEA 0183-compliant channels---wind (\texttt{MWV}), heading (\texttt{HDT}), and GPS-derived speed (\texttt{VTG})---records were filtered to entries carrying quality flag \texttt{A} (valid), removing samples affected by degraded satellite signals, checksum errors, or device-level faults.

Fuel-flow channels from the OpDAQ system were screened using embedded diagnostic flags reflecting flowmeter operational status; observations flagged as unreliable, typically during warm-up phases, communication dropouts, or calibration events, were excluded. This step prevented spurious or interpolated readings from biasing downstream model training and steady-state segmentation.

\subsubsection{Steady-State Cruising Data Extraction}
Representative cruising conditions were isolated to focus modeling on steady propulsion states. Observations with speed over ground (SOG) below 7~knots were removed to exclude low-speed operations such as maneuvering, docking, or ice navigation, which exhibit nonstationary fuel behavior.

Rotational motion was filtered using heading rate: heading values were differenced at 1~Hz to compute instantaneous rotation, and observations exceeding $\pm$5\textdegree/s were discarded, as such rates indicate sharp maneuvers and transient loading. This step filters on heading rate alone; rudder-induced corrections from wave action are not independently filtered here, but are caught by the subsequent steady-state detection step, which includes rudder angle (actual) as one of four monitored channels and gives a further check on directional stability.

A moving-window steady-state detection algorithm was then applied across four channels---fuel flow, SOG, rudder angle (actual), and heading---to identify periods of dynamic stability. Five-minute rolling averages were computed for each channel, and individual samples were validated against tolerance thresholds defined around the local rolling mean; thresholds were set independently per channel to reflect differing baseline variability. Samples exceeding their channel-specific tolerance were flagged as unstable.

Continuous stable intervals exceeding 20~minutes were identified, and a 2.5-minute buffer was trimmed from each end of every qualifying segment to remove edge effects at segment boundaries. Only the trimmed segments were retained.

Speed-through-water (STW) records were also screened for sensor dropout: samples with STW below 0.5~knots recorded concurrently with SOG above 7~knots were discarded, since this combination identifies signal-loss events at the relative-water sensor rather than genuine low-speed operation.

\subsubsection{Outlier Removal}
Following steady-state extraction, residual outliers were removed using the interquartile range (IQR) method, with samples beyond $1.5\times$IQR from the first and third quartiles discarded. This criterion was applied to four operational channels: main-engine fuel flow, speed over ground (SOG), apparent wind speed, and speed through water (STW).

This multi-stage refinement reduced the dataset from 26{,}082{,}125 raw observations to 19{,}494{,}959 quality-screened entries, and finally to 3{,}887{,}002 steady-state cruising records at 1~Hz.

\subsection{Feature Engineering}
A suite of derived features was constructed from onboard sensor measurements and hindcast environmental parameters to strengthen the model's predictive capability. These engineered variables represent vessel--environment interactions and operational states that directly influence fuel consumption.

Total fuel flow was calculated by summing the instantaneous flow rates from all three main engines, forming a unified target variable for model training. Directional parameters---including wind, wave, and swell---were converted into relative angles with respect to the ship's true heading. To capture qualitative directional effects, these relative angles were further discretized into eight categorical sectors of 45\textdegree\ each, as defined in Table~4.

Hydrodynamic influences were represented through a derived wave and swell height-period ratio, expressed as the ratio of significant height to mean period---an established proxy for sea-energy intensity and encounter severity, with units of m/s. Draft-related features were also included: fore and aft draft were obtained as continuous measurements from the same onboard sensor infrastructure used for SOG and engine performance channels, and the mean of the two gave a measure of vessel displacement, while their difference (trim) quantified longitudinal attitude and hull immersion balance.
Table~\ref{tab:engineered_features} summarizes the engineered features, their derivation, and associated units.

\begin{table}[htbp]
\centering
\caption{Engineered features derived from onboard sensor and ERA5 hindcast channels.}
\label{tab:engineered_features}
\begin{tabular}{l p{6.0cm} l}
\toprule
\textbf{Feature} & \textbf{Derivation} & \textbf{Unit} \\
\midrule
\texttt{Total\_ME\_Flow} & \texttt{ME1\_Flow + ME2\_Flow + ME3\_Flow} & L/h \\
\texttt{Relative\_wave\_direction} & (\texttt{mwd} $-$ \texttt{Heading} + 360) mod 360 & degrees \\
\texttt{Relative\_swell\_direction} & (\texttt{mdts} $-$ \texttt{Heading} + 360) mod 360 & degrees \\
\texttt{Wind\_Direction\_Class} & $\lfloor (\theta + 22.5)/45 \rfloor$ mod 8, $\theta = $ \texttt{MWV Wind Angle} & -- \\
\texttt{Wave\_Direction\_Class} & $\lfloor (\theta + 22.5)/45 \rfloor$ mod 8, $\theta = $ \texttt{Relative\_wave\_direction} & -- \\
\texttt{Swell\_Direction\_Class} & $\lfloor (\theta + 22.5)/45 \rfloor$ mod 8, $\theta = $ \texttt{Relative\_swell\_direction} & -- \\
\texttt{Wave Intensity Ratio} & \texttt{swh / mwp} & m/s \\
\texttt{Swell Intensity Ratio} & \texttt{shts / mpts} & m/s \\
\texttt{Mean\_Draft} & (\texttt{Fore Draft} + \texttt{Aft Draft}) / 2 & m \\
\texttt{Trim} & \texttt{Fore Draft} $-$ \texttt{Aft Draft} & m \\
\bottomrule
\end{tabular}
\end{table}

Engine RPM and thrust-related parameters are available from the vessel's propulsion control system but were deliberately excluded from the dataset used in this study. Both quantities are control-state outputs rather than planning-stage inputs: RPM and thrust are determined by the propulsion control system in response to a commanded speed under the prevailing resistance and loading conditions, and therefore already implicitly encode much of the fuel-flow response the model seeks to predict. Including such variables would risk a form of target leakage, inflating apparent predictive accuracy without providing information that is actually available ahead of a voyage. Since the intended application of this model is forward decision support for route and speed planning---where only planning-stage variables such as target speed, draft, and forecast environmental conditions are known in advance---RPM and thrust were left out so that the feature set remains representative of genuinely available planning-time information.

\subsection{Exploratory Data Analysis}

\begin{figure}[htbp]
\centering
\includegraphics[width=0.85\textwidth]{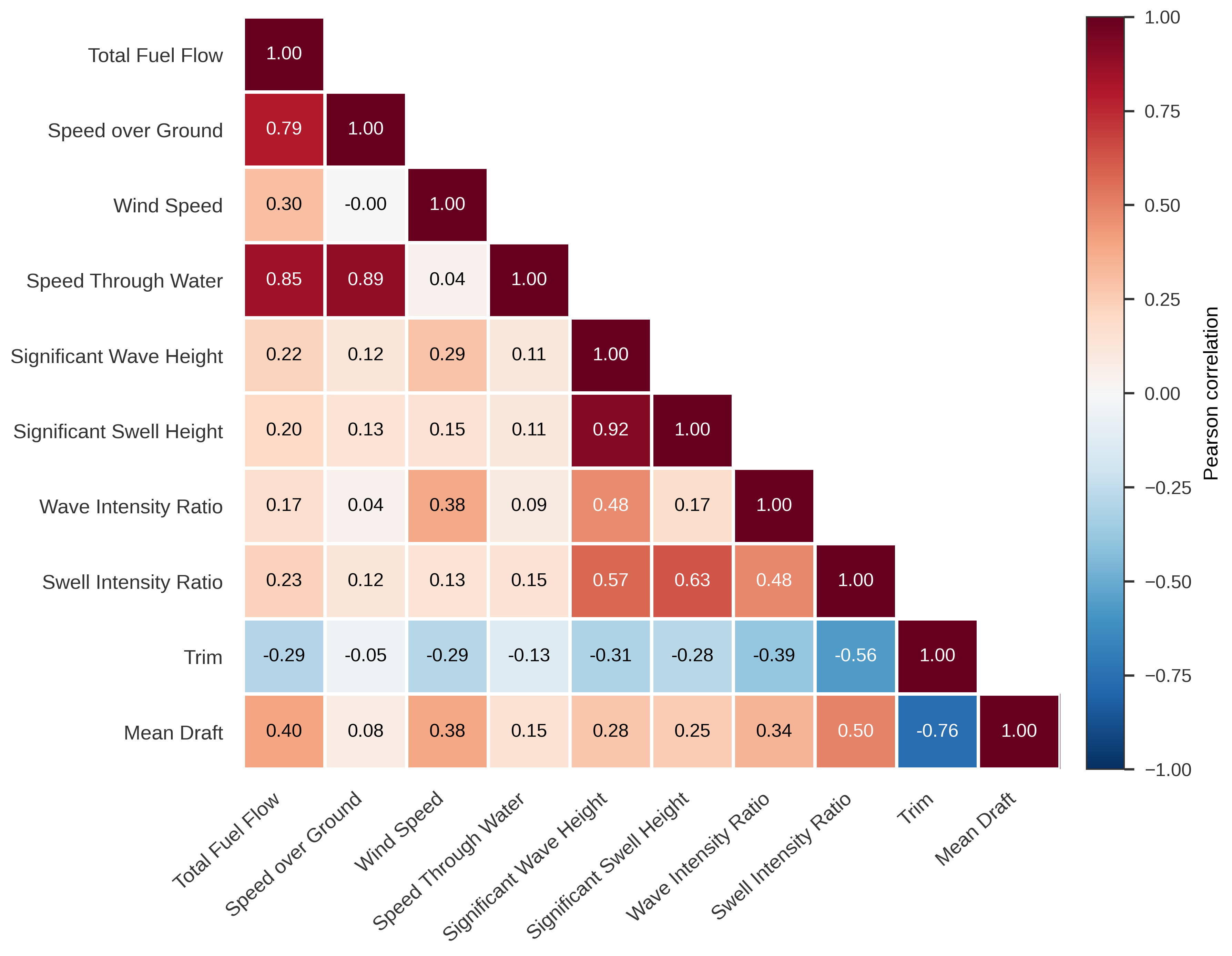}
\caption{Pearson correlation matrix of engineered features.}
\label{fig:corr_matrix}
\end{figure}

Exploratory data analysis was conducted on the cleaned, steady-state cruising dataset to characterize variable distributions and identify relationships among key predictors. Figure~\ref{fig:corr_matrix} presents the Pearson correlation matrix for the final set of engineered features. Total fuel flow shows a strong positive correlation with speed over ground ($r = 0.79$), consistent with fuel consumption scaling with speed under normal cruising conditions. Wave- and swell-related parameters show moderate mutual correlation---most notably between significant wave height and significant swell height ($r = 0.92$)---reflecting their shared dependence on ocean wave energy. The derived height-period ratio features correlate weakly with propulsion variables but moderately with draft and trim, suggesting a secondary hydrodynamic pathway. Overall, inter-predictor correlations remain low to moderate, indicating minimal risk of multicollinearity in the regression models developed in Section~4.

\begin{figure}[htbp]
\centering
\includegraphics[width=\textwidth]{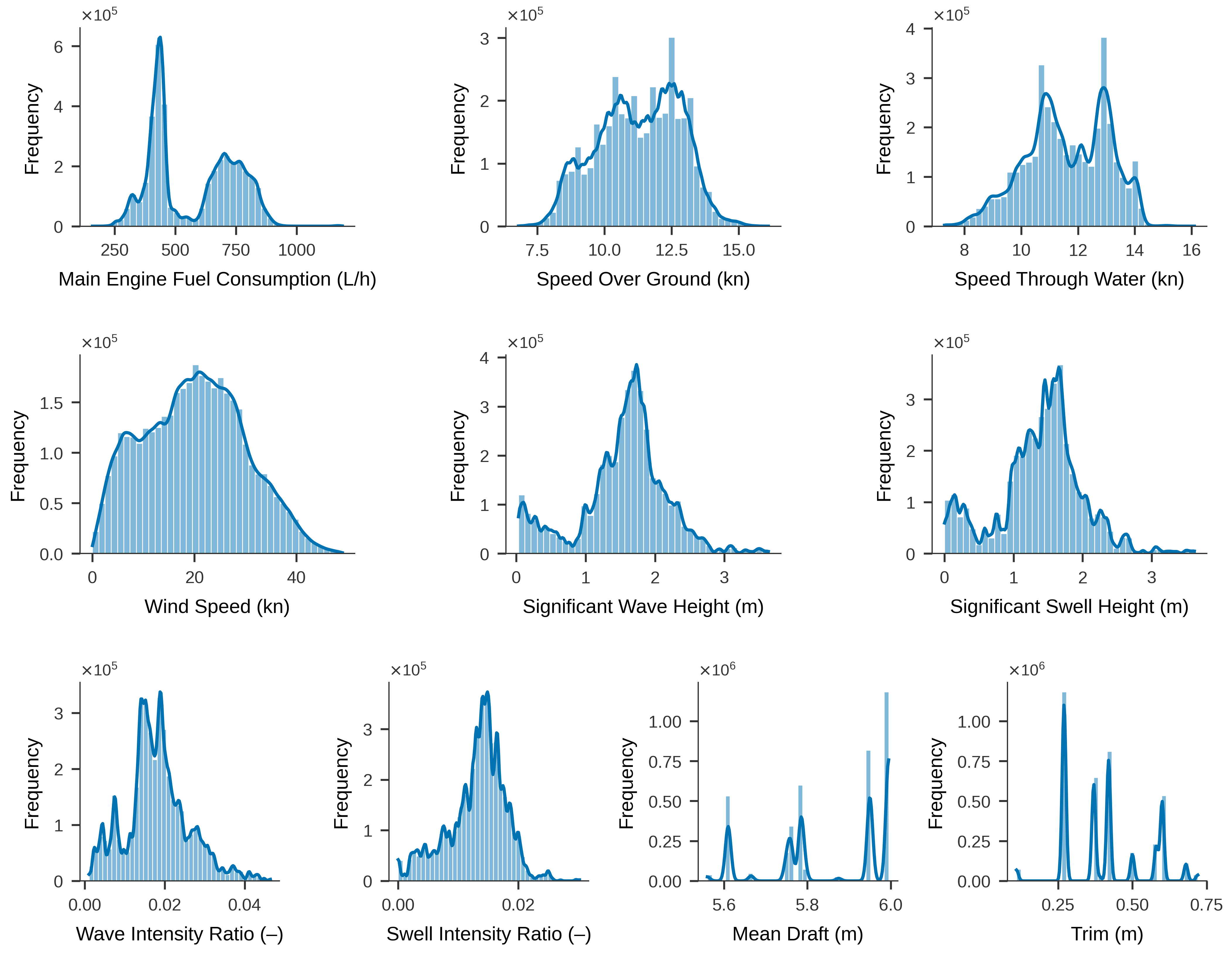}
\caption{Distributions of principal numerical features used for model training.}
\label{fig:eda_dist}
\end{figure}

Figure~\ref{fig:eda_dist} presents the distributions of the principal numerical features. Total fuel flow is multimodal, consistent with distinct engine-load configurations across operating regimes. Speed over ground follows an approximately normal distribution centered near 12~knots, representative of sustained open-water cruising. Wind speed and sea-state indicators (significant wave height, significant swell height) are right-skewed, reflecting a predominance of moderate sea states punctuated by high-energy events; the derived height-period ratio features show similarly long-tailed behavior. Draft and trim distributions are discrete, with peaks corresponding to recurring ballast or cargo configurations.

\begin{figure}[htbp]
\centering
\includegraphics[width=\textwidth]{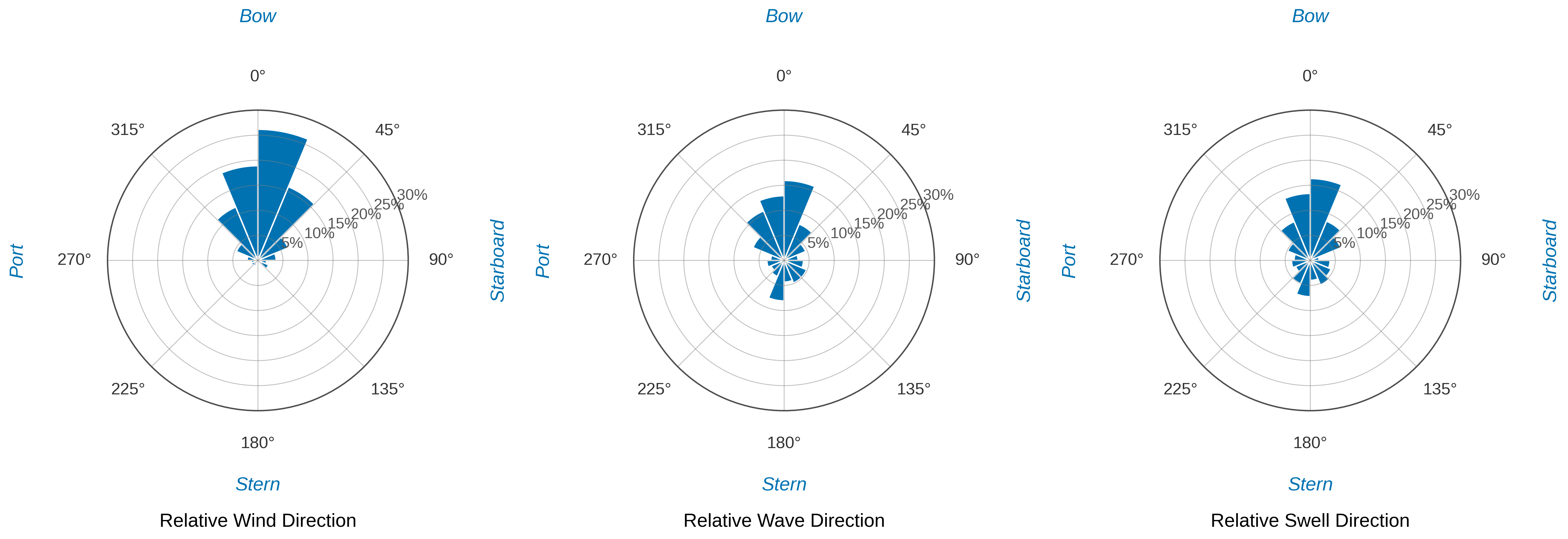}
\caption{Directional distributions of relative wind, wave, and swell encounter angles.}
\label{fig:direction_roses}
\end{figure}

Figure~\ref{fig:direction_roses} presents the directional distributions of relative wind, wave, and swell encounter angles. These distributions indicate the range of environmental headings encountered during the study period and motivate the directional binning scheme introduced in Section~3.3.

\section{Methodology}

\begin{figure}[p]
  \centering
  \includegraphics[height=0.92\textheight,keepaspectratio]{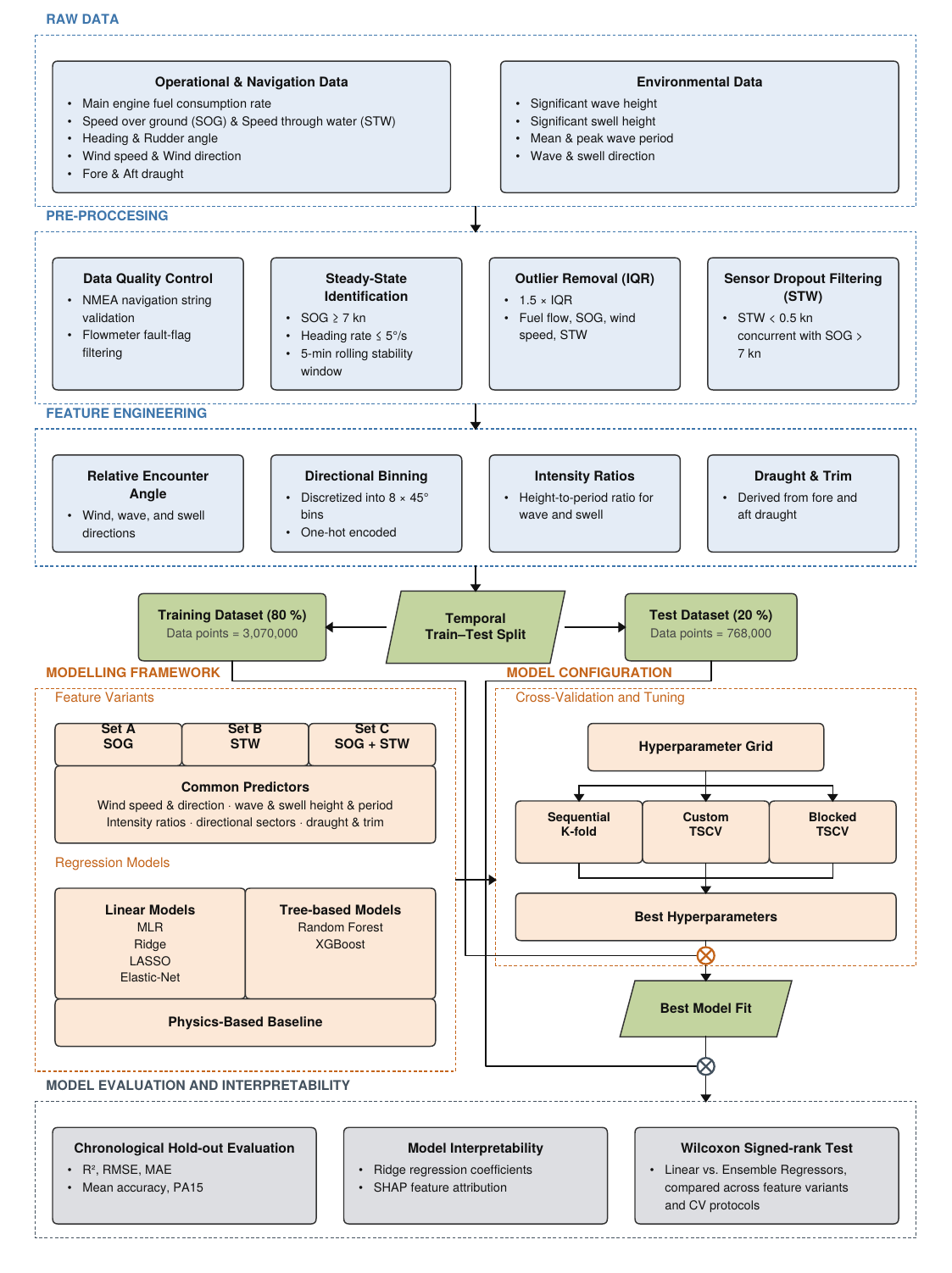}
  \caption{Overview of the modeling and validation pipeline, from raw sensor and
  hindcast inputs through preprocessing and feature engineering to model training,
  cross-validation, and final hold-out evaluation.}
  \label{fig:flow_chart}
\end{figure}

This section describes the modeling framework developed to estimate main-engine fuel consumption for the CCGS \textit{Sir Wilfrid Laurier}. Figure~\ref{fig:flow_chart} summarizes the complete pipeline, from raw sensor and hindcast inputs through preprocessing, feature engineering, and the chronological train--test partition, to the modeling framework, cross-validation and tuning stage, and the final evaluation and interpretability analyses. Fuel consumption is modeled as a supervised regression problem, with total main-engine fuel flow (L\,h$^{-1}$) as the response and operational and environmental variables derived from onboard sensors and ERA5 hindcast fields as predictors. The section defines the feature variants and regression models used, introduces a physics-based reference model benchmarked under identical conditions, and details the time-aware validation schemes, evaluation metrics, and hyperparameter tuning and significance-testing procedures adopted throughout the study.

\subsection{Problem Formulation and Feature Variants}

Let \( y_t \in \mathbb{R} \) denote total main-engine fuel flow (L\,h$^{-1}$) recorded at time \( t \), and let \( \mathbf{x}_t \in \mathbb{R}^{d} \) denote the corresponding vector of operational and environmental predictors. The objective is to estimate a mapping \( f: \mathbb{R}^{d} \rightarrow \mathbb{R} \) such that

\[
\hat{y}_t = f(\mathbf{x}_t),
\]

with \( f \) learned from the steady-state cruising records described in Section~3.2. No lagged or autoregressive terms derived from past fuel-flow measurements are included in \( \mathbf{x}_t \); each estimate depends only on conditions available at the time it describes, since a voyage-planning decision support system must estimate the fuel rate for a leg before it is sailed, when no fuel-flow measurements for that leg yet exist.

Three feature variants are defined, differing only in the vessel speed channel supplied to the model: Variant~A uses speed over ground (SOG) alone, Variant~B uses speed through water (STW) alone, and Variant~C supplies both. All three share a common predictor block of environmental, directional, and vessel-state features, listed in Table~4: wind speed and relative wind angle, significant wave and swell height, the wave and swell intensity ratios, relative wave and swell encounter angle (each discretized into eight $45^{\circ}$ sectors and one-hot encoded), and mean draft and trim. Table~5 defines the three variants and the question each is intended to answer.

\begin{table}[htbp]
\centering
\caption{Feature variant definitions. All variants share the same environmental,
directional, and draft-related features (Table~4) and differ only in the vessel
speed channel used.}
\label{tab:variant_defs}
\begin{tabular}{@{}c l p{7.2cm}@{}}
\toprule
\textbf{Variant} & \textbf{Speed Channel} & \textbf{Purpose} \\
\midrule
\textbf{A} & Speed over Ground (SOG) &
Available at the voyage-planning stage; the configuration evaluated for deployment. \\[4pt]
\textbf{B} & Speed through Water (STW) &
Relates fuel consumption directly to hull resistance, independent of surface current and drift. \\[4pt]
\textbf{C} & SOG + STW &
Tests whether combining both signals improves accuracy beyond either alone, using their difference as an implicit proxy for current and drift. \\
\bottomrule
\end{tabular}
\end{table}

The two speed signals are not interchangeable, and the choice of which to include is the substantive difference between variants. SOG, obtained from the vessel's GPS, is the speed a planner specifies before a voyage begins, making it the only speed signal genuinely available at the planning stage. STW, obtained from the Doppler log, is the water-relative velocity and the physically appropriate argument for hull resistance, but it cannot be specified ahead of a voyage without a forecast of surface currents, and is subject to the log-dropout behavior described in Section~3.2. Variant~C supplies both signals, letting the models exploit their difference as an implicit proxy for current and drift. Variant~A is therefore the configuration evaluated for deployment, while Variants~B and~C serve as reference points that isolate the contribution of each speed channel.

\subsection{Regression Models}

Six regression models were evaluated, spanning two families: penalized linear estimators with closed-form or convex solutions, and nonlinear tree-based ensembles fitted by iterative optimization. The two families make different assumptions about the form of the speed--power relationship and about how environmental predictors combine, and they differ correspondingly in their capacity to fit and to overfit. The mathematical foundations of each are summarized below.

\paragraph{Multiple Linear Regression (MLR)}
Multiple Linear Regression assumes a linear mapping between the predictor matrix
\( X = [x_1, x_2, \ldots, x_p] \) and the response \( y \). The parameters
\( \boldsymbol{\beta} \) are estimated by minimizing the residual sum of
squares~\cite{Hastie2009ESL}:

\[
\hat{y} = \beta_0 + \sum_{j=1}^{p} \beta_j x_j = X\boldsymbol{\beta},
\qquad
\hat{\boldsymbol{\beta}} = \arg\min_{\boldsymbol{\beta}} \, \| y - X\boldsymbol{\beta} \|_2^2
\]

MLR provides direct interpretability through explicit coefficient estimates, but offers no mechanism for controlling variance when predictors are correlated. It is retained here as an unregularized reference against which the three penalized linear models can be read.

\paragraph{Ridge Regression}
Ridge Regression augments the least-squares objective with an \( L_2 \) penalty on coefficient magnitude, shrinking estimates toward zero and stabilizing them under predictor correlation~\cite{hoerl1970ridge}:

\[
\hat{\boldsymbol{\beta}}^{\,\mathrm{ridge}} = \arg\min_{\boldsymbol{\beta}}
\left\{ \| y - X\boldsymbol{\beta} \|_2^2 + \alpha \|\boldsymbol{\beta}\|_2^2 \right\}
\]

where \( \alpha > 0 \) controls the degree of shrinkage. The penalty introduces bias in exchange for a reduction in variance, and retains all predictors in the model rather than selecting among them---a property that suits a feature set in which the environmental and directional variables are individually weak but jointly informative.

\paragraph{Lasso Regression}
Lasso (Least Absolute Shrinkage and Selection Operator) replaces the \( L_2 \) penalty with an \( L_1 \) penalty, which drives a subset of coefficients exactly to zero and thereby performs embedded feature selection~\cite{tibshirani1996regression}:

\[
\hat{\boldsymbol{\beta}}^{\,\mathrm{lasso}} = \arg\min_{\boldsymbol{\beta}}
\left\{ \| y - X\boldsymbol{\beta} \|_2^2 + \alpha \|\boldsymbol{\beta}\|_1 \right\}
\]

The resulting model is more compact than the Ridge solution, at the cost of instability in which predictor is retained when several are strongly correlated.

\paragraph{ElasticNet Regression}
ElasticNet combines both penalties, balancing coefficient shrinkage against variable selection~\cite{zou2005regularization}:

\[
\hat{\boldsymbol{\beta}}^{\,\mathrm{enet}} = \arg\min_{\boldsymbol{\beta}}
\left\{
\| y - X\boldsymbol{\beta} \|_2^2
+ \alpha \left[ \rho \|\boldsymbol{\beta}\|_1 + (1-\rho)\|\boldsymbol{\beta}\|_2^2 \right]
\right\}
\]

where \( \rho \in [0,1] \) is the mixing parameter governing the trade-off between the two terms, and \( \rho = 1 \) and \( \rho = 0 \) recover Lasso and Ridge respectively. ElasticNet is effective when predictors exhibit grouped correlation, as is common among environmental and hydrodynamic variables, because correlated predictors tend to be retained or discarded together rather than arbitrarily.

\paragraph{Random Forest Regression}
Random Forest (RF) is a bagging ensemble that fits \( B \) regression trees to bootstrap replicates of the training data and averages their predictions~\cite{breiman2001random}:

\[
\hat{y} = \frac{1}{B} \sum_{b=1}^{B} f_b(x)
\]

where \( f_b \) denotes the tree fitted to the \( b \)-th replicate. Each tree recursively partitions the feature space to minimize node impurity, measured here by mean squared error, and a random subset of features is considered at every split. Bootstrap resampling and feature sub-sampling together decorrelate the individual trees, so that averaging reduces variance without a corresponding increase in bias. Random Forests capture nonlinear effects and predictor interactions without requiring these to be specified in advance, but their predictions are bounded by the range of response values observed during training.

\paragraph{Extreme Gradient Boosting (XGBoost)}
XGBoost constructs an additive ensemble of regression trees in a stage-wise manner, each fitted to the residual signal left by its predecessors~\cite{chen2016xgboost}. At boosting iteration \( k \), the prediction is updated as

\[
\hat{y}_i^{(k)} = \hat{y}_i^{(k-1)} + \eta \, f_k(x_i),
\qquad f_k \in \mathcal{F}
\]

where \( \mathcal{F} \) is the space of regression trees and \( \eta \) is the learning rate. The tree \( f_k \) is selected to minimize the regularized objective

\[
\mathcal{L}^{(k)} =
\sum_{i=1}^{n} l\!\left(y_i,\; \hat{y}_i^{(k-1)} + f_k(x_i)\right)
+ \Omega(f_k),
\qquad
\Omega(f) = \gamma T + \tfrac{1}{2}\lambda_2 \|w\|_2^2 + \lambda_1 \|w\|_1
\]

in which \( l(\cdot) \) is a differentiable loss function (squared error throughout this study), \( T \) is the number of leaves in the tree, \( w \) is the vector of leaf weights, and \( \gamma \), \( \lambda_1 \), and \( \lambda_2 \) penalize tree complexity and leaf-weight magnitude. XGBoost extends conventional gradient boosting through a second-order approximation of the loss, shrinkage via \( \eta \), and column and row subsampling. Its flexibility makes it well suited to nonlinear dependencies and interaction effects in high-frequency operational data.

All six models are fitted through the common preprocessing pipeline described invSection~4.1, applied identically across every feature variant. MLR has no tunablevhyperparameters and is therefore carried through unchanged, while the remaining five are tuned as described in Section~4.6.

\subsection{Physics-Based Baseline}
A physics-based reference model was included alongside the six regression models, derived from the classical resistance relations reviewed in Section~2.1.1. Frictional and residuary resistance both scale approximately with the square of vessel speed, so propulsion power, being the product of resistance and speed, scales with the cube; fuel flow at a fixed engine efficiency is proportional to delivered power, giving a leading cubic term in speed. Mean draft is included as a linear term, serving as the available proxy for loading condition. The resulting specification is
\[
\widehat{FC} = a\,v^{3} + c\,d + b
\]
where \( FC \) is total main-engine fuel flow (L\,h$^{-1}$), \( v \) is speed over ground (kn), and \( d \) is mean draft (m). Environmental forcing is not included; the baseline represents the speed--displacement relationship underlying early-phase powering estimates, not a complete resistance model.

The model uses Variant~A (speed over ground) only, and its three coefficients were estimated by ordinary least squares on the training partition. Fitted on the full training set, the calibration is
\[
\widehat{FC} = 0.2184\,v^{3} + 402.97\,d - 2118.97
\]
Figure~5 shows the corresponding response surface over the observed operating range. The surface rises steeply with speed and shifts upward with increasing draft, consistent with expected vessel behavior.

\begin{figure}[htbp]
  \centering
  \includegraphics[width=\textwidth]{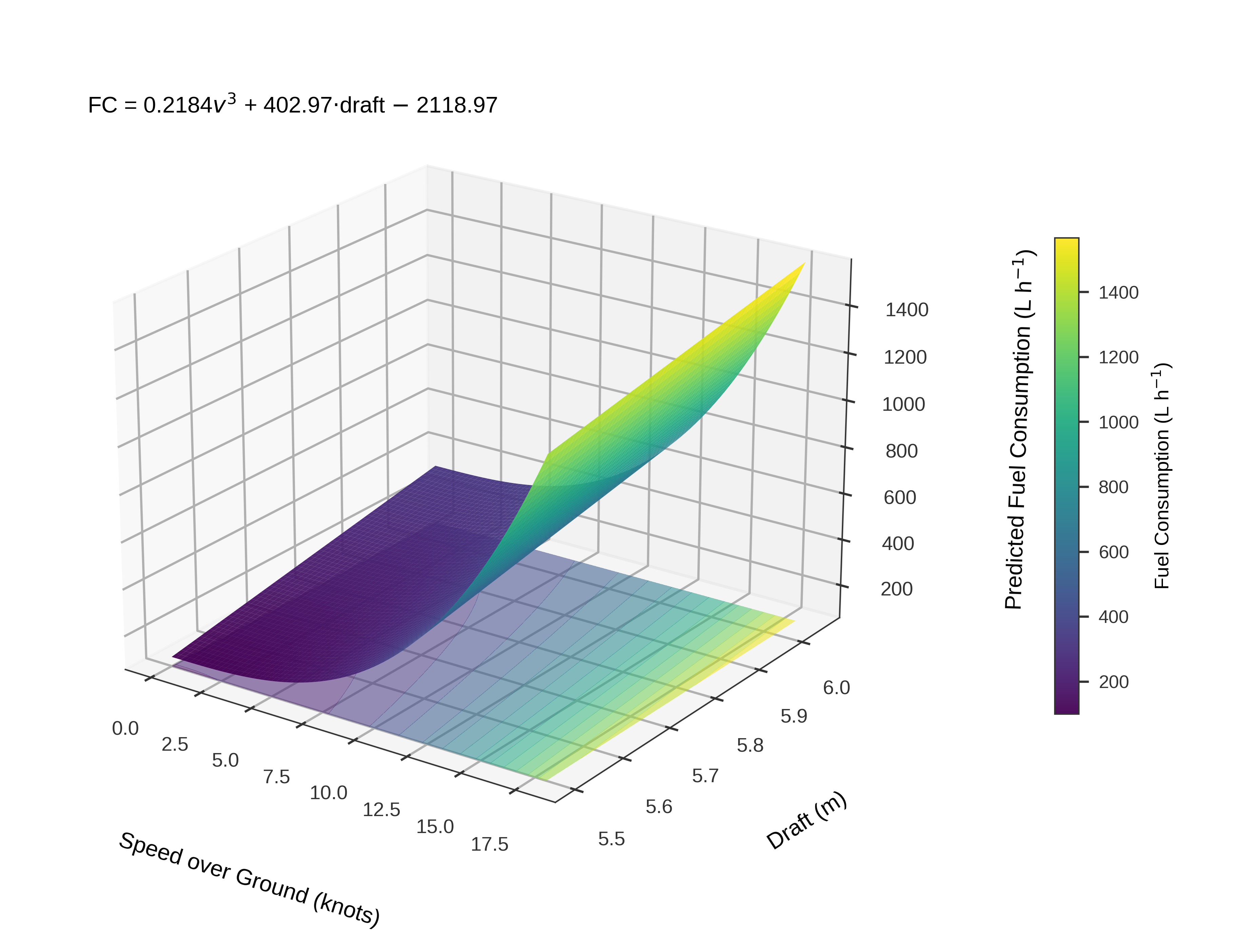}
  \caption{Physics baseline model fitted using OLS: predicted fuel flow as a function of speed over ground and mean draft.}
\end{figure}

\subsection{Validation Strategy}
\subsubsection{Importance of Proper Validation}

Validation determines how well a predictive model is expected to perform beyond the data on which it was trained, and the way the data are partitioned directly shapes that measurement. In ship fuel-consumption modeling, random 70:30 or 80:20 train--test splits and standard $k$-fold cross-validation are the most widely used strategies, and as summarized in Table~1 they appear across much of the recent literature. These methods remain widely used because they are simple, computationally efficient, and statistically sound whenever observations can be treated as independent and identically distributed. That independence assumption is only partially met by high-frequency vessel records. Ship dynamics, engine loading, and environmental forcing all evolve gradually, so consecutive 1\,Hz observations are strongly autocorrelated and adjacent samples are close to duplicates of one another. Under random partitioning, the samples immediately preceding and following any validation observation are almost certain to have been placed in the training set. The model is then evaluated on states it has effectively already seen, and the resulting score measures interpolation within a known operating window rather than generalization to an unseen one.

The severity of this effect scales with sampling rate. Coarse-resolution records such as daily noon reports are separated by long intervals and are comparatively little affected, which is one reason random splitting has served that literature well. At 1\,Hz, by contrast, even a small overlap between adjacent folds is enough to bias the outcome, and the bias acts in a specific direction: it rewards models flexible enough to memorize local structure, since that structure is present on both sides of the split. Evaluation that reflects deployment therefore requires partitions that preserve chronological order and enforce a clean separation between past and future.

Three validation schemes are adopted in this study, forming a graded sequence in how strictly temporal separation is enforced. Each is illustrated below using a small example of 20 samples divided into 5 folds, so that the resulting fold structure is easy to see; the fold counts and split ratios actually used follow in the corresponding subsection.

\subsubsection{Sequential K-Fold Cross-Validation}

\begin{figure}[htbp]
  \centering
  \includegraphics[width=\textwidth]{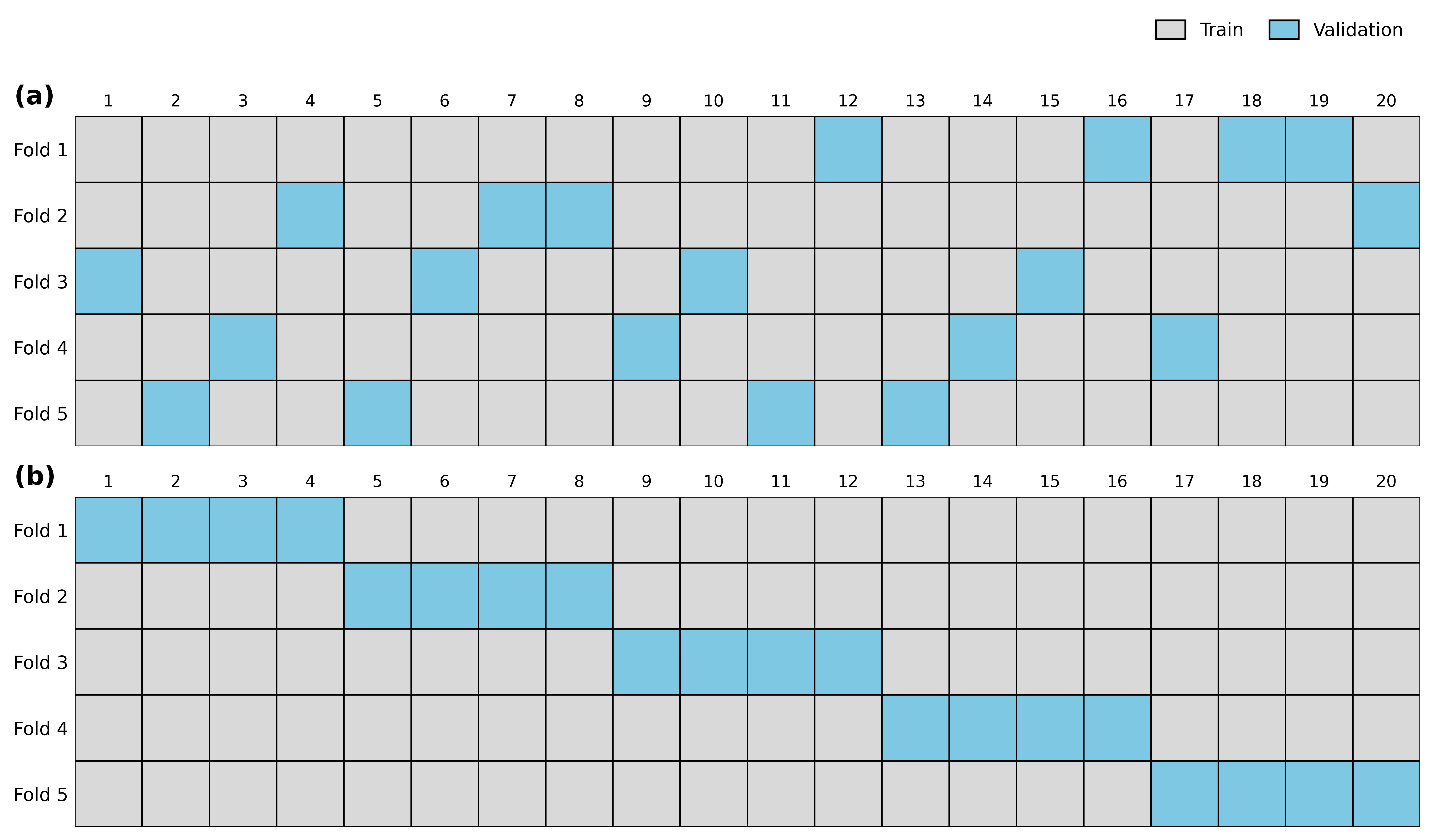}
  \caption{Cross-validation fold structure for (a) random $k$-fold and
  (b) sequential $k$-fold, shown for $n = 20$ samples and $k = 5$ folds.}
  \label{fig:kfold_pair}
\end{figure}

The first scheme retains the conventional $k$-fold structure but removes shuffling. The dataset is divided into $k$ contiguous, equally sized segments in chronological order, and each segment serves in turn as the validation fold while the remaining $k-1$ segments form the training set. Five folds were used ($k=5$). Figure~\ref{fig:kfold_pair} compares this arrangement with random $k$-fold partitioning of the same records.

Under random $k$-fold (Figure~\ref{fig:kfold_pair}a), validation samples are scattered throughout the record, so most validation points have neighboring training samples on either side. Under sequential $k$-fold (Figure~\ref{fig:kfold_pair}b), each validation fold forms a single contiguous block, and validation samples are drawn from a period of the record distinct from their corresponding training data.

Sequential $k$-fold is nonetheless only partially time-aware. Because every segment other than the validation fold is used for training, folds other than the last are still fitted on data that includes observations occurring after the validation period. Chronological order is preserved within each fold, but training is not restricted to the past relative to validation, as it would be in genuine forecasting. Sequential $k$-fold is treated here as the minimally time-aware baseline against which the two stricter schemes below are compared.

\subsubsection{Time Series Cross-Validation}

Time series cross-validation (TSCV) enforces the forward-only constraint that sequential $k$-fold does not. In its standard, expanding-window form~\cite{bergmeir2012crossvalidation}, the training window grows forward in time and validation is performed on the block immediately following it. For a dataset $\{(\mathbf{x}_t, y_t)\}_{t=1}^{T}$, the $i$-th fold is defined as

\[
\mathcal{D}^{\mathrm{train}}_i = \{(\mathbf{x}_t, y_t) \mid 1 \leq t \leq t_i\},
\qquad
\mathcal{D}^{\mathrm{val}}_i = \{(\mathbf{x}_t, y_t) \mid t_i < t \leq t_{i+1}\},
\]

with $t_i < t_{i+1}$, so that no validation observation precedes any observation used to train the corresponding fold.

\begin{figure}[htbp]
  \centering
  \includegraphics[width=\textwidth]{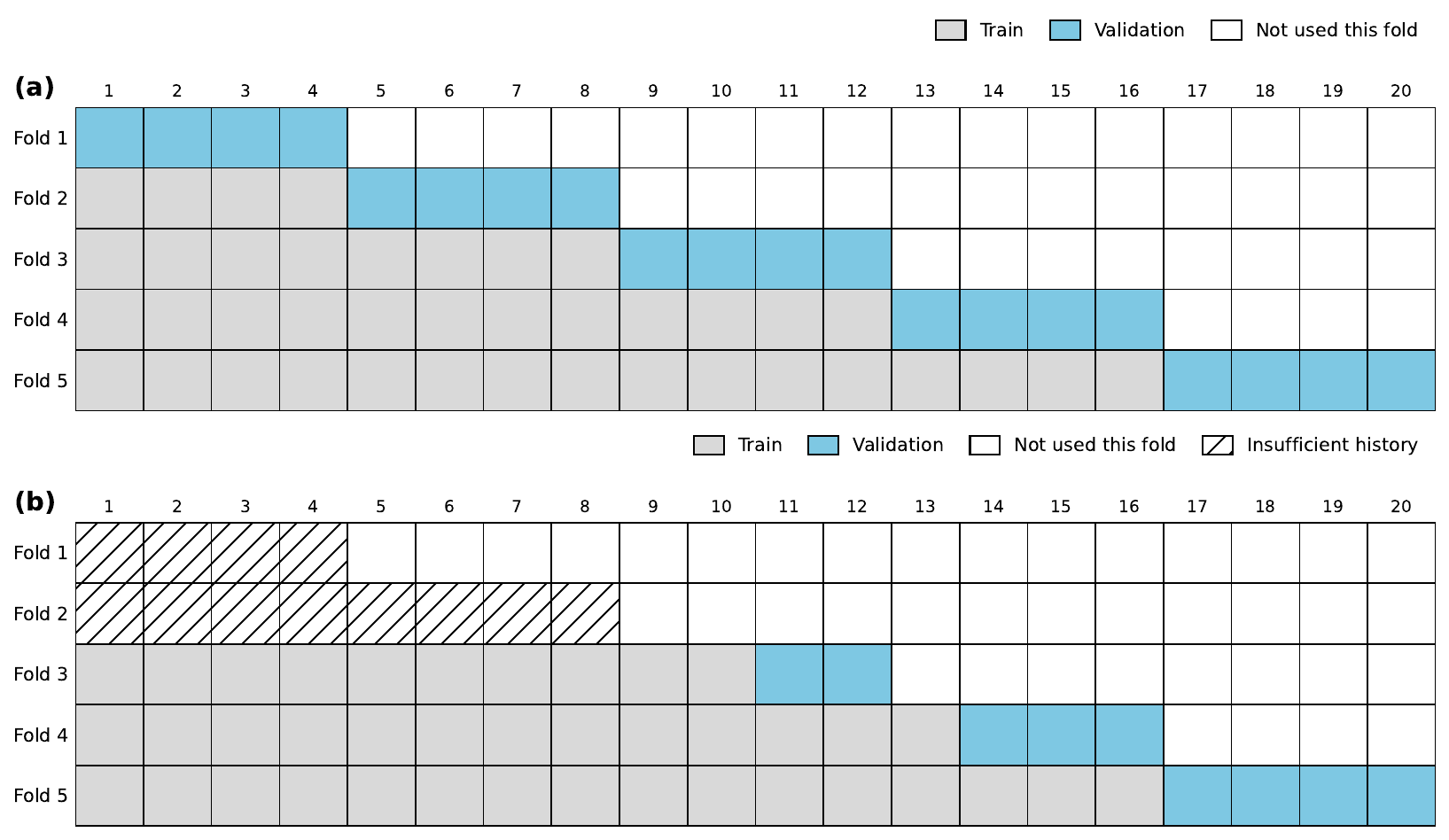}
  \caption{Time series cross-validation fold structure for (a) the standard
  expanding-window formulation and (b) the modified scheme adopted in this study,
  shown for 20 samples divided into 5 partitions.}
  \label{fig:tscv_pair}
\end{figure}

This formulation preserves chronological order, but the resulting folds are not comparable to one another. Because the validation block has a fixed size while the training window grows, the ratio of training to validation data changes from fold to fold: early folds have very little training history, and late folds reserve only a small fraction of the record for validation (Figure~\ref{fig:tscv_pair}a). Averaging performance across folds constructed this way combines estimates obtained under different conditions, which makes the resulting mean difficult to interpret.

A modified expanding-window scheme was adopted to address this. The record is divided into $k$ equal temporal partitions, and the earliest partitions are excluded from evaluation because they provide insufficient training history. From the first evaluated partition onward, all preceding data are used for training, and a fixed 80:20 chronological split is applied within each cumulative window. Ten partitions were used ($k = 10$), with the first five excluded and folds 6--10 evaluated.

Figure~\ref{fig:tscv_pair}b shows the result. Every evaluated fold has the same 80:20 training-to-validation ratio, so performance can be averaged across folds without combining estimates from differently sized windows. The forward-only property of the standard formulation is retained: within each fold, every validation sample occurs after every training sample.

\subsubsection{Blocked Time Series Validation}

\begin{figure}[htbp]
  \centering
  \includegraphics[width=\textwidth]{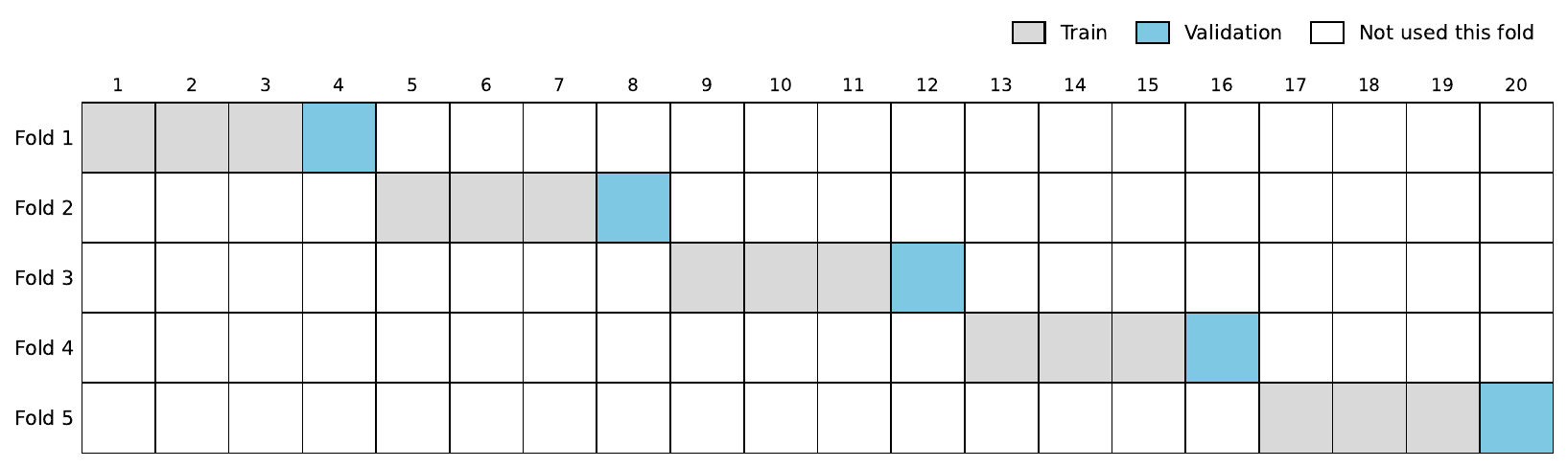}
  \caption{Blocked time series validation fold structure, shown for 20 samples
  divided into 5 blocks.}
  \label{fig:blocked_tscv}
\end{figure}

The third scheme replaces the cumulative training window with independent temporal blocks. The record is partitioned into $k$ non-overlapping contiguous blocks of equal length, and within each block the earliest portion is used for training and the final portion for validation. Three blocks were used ($k = 3$), with a 90:10 training-to-validation split applied within each. Figure~\ref{fig:blocked_tscv} illustrates the arrangement.

Blocked validation differs from the two expanding-window schemes in that no data are shared between folds. Under standard or modified TSCV, every fold is trained on the earliest portion of the record, so the folds are nested and their performance estimates are not independent of one another. Under blocked validation, each fold is trained and evaluated using only its own block, with no information carried over from any other segment. This makes the scheme well suited to records that span distinct operational regimes---separate voyages, missions, or seasonal conditions---since it measures how a model performs when trained and evaluated entirely within one such regime. The spread of results across blocks then reflects consistency across regimes, rather than the effect of accumulating more training data over time.

\subsection{Evaluation Metrics}
Model performance was assessed using five metrics, each capturing a different aspect of predictive accuracy relevant to the deployment context of a voyage-planning decision support system.
\paragraph{Root Mean Squared Error (RMSE)}
RMSE measures the square root of the average squared difference between predicted and observed fuel flow, and penalizes large errors more heavily than small ones~\cite{Chai2014RMSE}:
\[
\mathrm{RMSE} = \sqrt{\frac{1}{n}\sum_{i=1}^{n}(y_i - \hat{y}_i)^2}
\]
\paragraph{Mean Absolute Error (MAE)}
MAE reports the average magnitude of prediction error in the original units of fuel flow (L\,h$^{-1}$), without disproportionately weighting large deviations~\cite{Chai2014RMSE}:
\[
\mathrm{MAE} = \frac{1}{n}\sum_{i=1}^{n} |y_i - \hat{y}_i|
\]
\paragraph{Coefficient of Determination ($R^2$)} $R^2$ expresses the proportion of variance in fuel flow explained by the model, relative to a constant-mean predictor~\cite{Chai2014RMSE}:
\[
R^2 = 1 - \frac{\sum_i (y_i - \hat{y}_i)^2}{\sum_i (y_i - \bar{y})^2}
\]
\paragraph{Mean Accuracy}
Mean accuracy converts the average relative error into a single interpretable percentage~\cite{ZHOU2023115509}:
\[
\mathrm{Mean\ Accuracy} = \left(1 - \frac{1}{n}\sum_{i=1}^{n}
\left|\frac{y_i - \hat{y}_i}{y_i}\right|\right) \times 100
\]
\paragraph{Prediction Accuracy within 15\% (PA15)} PA15 reports the percentage of predictions falling within $\pm 15\,\%$ of the observed value~\cite{ZHOU2023115509}:
\[
\mathrm{PA15} = \frac{100}{n}\sum_{i=1}^{n}
\mathbb{1}\!\left(\left|\frac{\hat{y}_i - y_i}{y_i}\right| \leq 0.15\right)
\]
PA15 is reported alongside RMSE, MAE, $R^2$, and Mean Accuracy because it expresses model performance as a bounded, interpretable error margin rather than as a variance-relative or unit-scaled statistic, and is therefore the metric most directly interpretable in terms of operational deployment.

\subsection{Hyperparameter Tuning}
Hyperparameters were tuned independently for each combination of feature variant, cross-validation scheme, and tunable model, giving 45 tuning runs in total (3 variants $\times$ 3 CV schemes $\times$ 5 models). MLR has no tunable hyperparameters and was excluded from this search. For every run, a randomized search was performed over the corresponding model's candidate grid, using the precomputed folds of the assigned cross-validation scheme and scoring each candidate by validation root-mean-square error; the configuration with the lowest validation RMSE was retained and refit on the full training partition for that fold structure. Table~6 lists the full search space and the number of search iterations used for each model; for Ridge and Lasso, the number of iterations equals the number of grid points, so the search evaluates the entire grid rather than a subsample of it. Across the 45 runs, this amounts to 1{,}125 candidate configurations evaluated in total, and 4{,}875 individual model fits once the folds of each cross-validation scheme are accounted for.

Random Forest was tuned on a representative 300{,}000-row subsample of the training data for computational tractability; all other models were tuned on the full training set for the corresponding feature variant. This reduction applies only to the search stage: the selected Random Forest configuration was refit on the full
training partition, consistent with every other model.

\begin{table}[htbp]
\centering
\footnotesize
\caption{Hyperparameter search space and tuning budget for each regression model.}
\label{tab:hyperparam_search_space}
\begin{threeparttable}
\begin{tabular}{l p{6.8cm} p{6cm} p{1cm}}
\toprule
\textbf{Model} & \textbf{Hyperparameter} & \textbf{Search Grid} & \textbf{Search Iterations} \\
\midrule
Ridge & Regularization strength ($\alpha$) & 30-point logarithmic grid, $10^{-3}$ to $10^{3}$ & 30\tnote{a} \\
\midrule
Lasso & Regularization strength ($\alpha$) & 20-point logarithmic grid, $10^{-3}$ to $10^{1}$ & 20\tnote{a} \\
\midrule
\multirow{2}{*}{ElasticNet} & Regularization strength ($\alpha$) & 15-point logarithmic grid, $10^{-2}$ to $10^{1}$ & \multirow{2}{*}{30} \\
 & Mixing parameter ($\ell_1$ ratio) & 9-point linear grid, 0.1 to 0.9 & \\
\midrule
\multirow{5}{*}{Random Forest} & Number of trees & 50, 100, 150, 200 & \multirow{5}{*}{20} \\
 & Maximum tree depth & 10, 20, 30 & \\
 & Minimum samples required to split a node & 2, 5, 10 & \\
 & Minimum samples required at a leaf & 1, 2, 4 & \\
 & Features considered per split & $\sqrt{p}$, $\log_2 p$, or all features & \\
\midrule
\multirow{9}{*}{XGBoost} & Number of trees & 100, 200, 321, 400, 600 & \multirow{9}{*}{25} \\
 & Maximum tree depth & 3, 5, 7, 9 & \\
 & Learning rate & 0.01, 0.05, 0.10, 0.15, 0.20 & \\
 & Subsample ratio (instances) & 0.7, 0.8, 0.9, 1.0 & \\
 & Subsample ratio (features) & 0.6, 0.75, 0.9, 1.0 & \\
 & Minimum child weight & 1, 3, 5, 7 & \\
 & Minimum loss reduction for a split ($\gamma$) & 0, 1, 2, 4.75 & \\
 & $L_1$ regularization coefficient & 0.0, 0.05, 0.15, 0.5 & \\
 & $L_2$ regularization coefficient & 0.5, 0.6, 1.0, 2.0 & \\
\bottomrule
\end{tabular}
\begin{tablenotes}
\small
\item[a] The full search grid was explored for this model.
\end{tablenotes}
\end{threeparttable}
\end{table}

\subsection{Statistical Significance and Model Explainability}

\subsubsection{Statistical Significance Testing}

Performance differences observed on the chronological hold-out set were tested for statistical significance rather than reported as point estimates alone. Two comparisons were of primary interest: whether the best-performing model differed significantly from the worst-performing model, and whether the tree-based ensembles differed significantly from the linear family as a whole. For a given pair of models, the loss differential at each test-set observation was defined as

\[
d_t = (y_t - \hat{y}_{1,t})^2 - (y_t - \hat{y}_{2,t})^2,
\]

the difference in squared error between the two models at time $t$. Each comparison was evaluated separately for every combination of feature variant and cross-validation scheme, so that any significant difference could be attributed to a specific validation and feature setting rather than obscured by pooling across
them.

Observations of $d_t$ are not independent, since the underlying record is sampled at 1\,Hz and adjacent rows reflect nearly identical operating conditions; a test that treated every row as an independent observation would substantially overstate the effective sample size. To account for this, the chronological test set was
divided into 100 contiguous, equally sized temporal blocks, and the mean loss differential was computed within each block. A Wilcoxon signed-rank test~\cite{wilcoxon1945individual} was then applied to the resulting 100 block means, testing the null hypothesis that they are symmetrically distributed about zero. This block-mean formulation requires only that the 100 block means be approximately independent of one another, a substantially weaker assumption than treating every individual row as independent, and is therefore better suited to strongly autocorrelated operational data. Because multiple comparisons were conducted across feature variants and validation schemes, a Benjamini--Hochberg false discovery rate correction~\cite{benjamini1995fdr} was applied across the full batch at $\alpha = 0.05$.

\subsubsection{Model Explainability}

To interpret what the models had learned, a representative linear model and the tree-based ensembles were refit on the complete training partition and examined using interpretability techniques appropriate to their respective structures. For the linear model, coefficients were extracted directly from the fit and compared in sign and relative magnitude, providing a globally consistent measure of feature influence under an additive, linear specification. For the tree-based ensembles, SHAP (SHapley Additive exPlanations) values were computed to quantify the marginal contribution of each feature to individual predictions~\cite{lundberg2017shap}, without imposing a linear or additive functional form on the underlying relationships. This division---coefficients for the linear model, SHAP values for the tree-based models---allows each family to be interpreted according to its own structure, while a normalized comparison across the resulting attributions places them on a common scale so that feature rankings remain comparable across model families.

\section{Results}

This section reports the experimental evaluation of the modeling framework introduced in Section 4. The evaluation is organised around a single methodological question: how sensitive is the apparent performance of fuel-consumption models to the design of the validation protocol used to assess them. We first contrast random and chronological data partitioning to establish the extent of this sensitivity, then examine performance under the three time-aware cross-validation schemes and across the three feature variants defined in Section 4.1. The final hold-out evaluation, formal significance testing, and model interpretability analyses follow, giving both a statistically grounded comparison of model performance and a physical interpretation of the underlying predictive relationships.

Condensed results are presented throughout the main text; the corresponding full metric sets — R², RMSE, MAE, mean accuracy, and PA15, reported for both training and evaluation partitions — are provided in Appendix A.

\subsection{Why Partitioning Strategy Matters: Random versus Time-Aware Splits}

\begin{figure}[p]
    \centering
    \includegraphics[width=\textwidth, height=0.9\textheight, keepaspectratio]{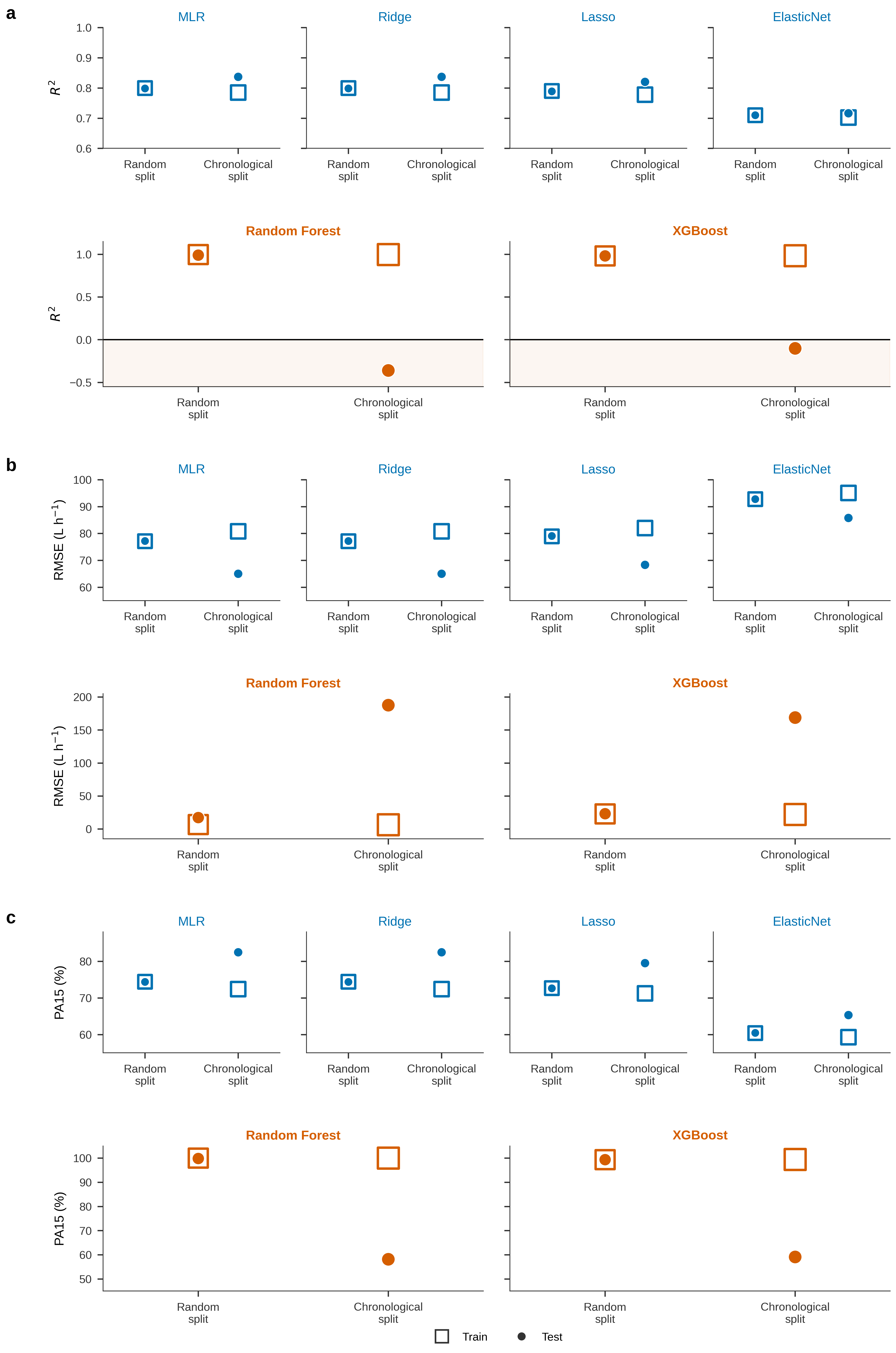}
    \caption{Model performance under random versus chronological 80:20 partitioning of Variant A data, reported as (a)~R\textsuperscript{2}, (b)~RMSE~(L~h\textsuperscript{-1}), and (c)~PA15~(\%).}
    \label{fig:train_test_random_vs_chronological}
\end{figure}

For fuel-consumption prediction, the choice of data-partitioning strategy can matter as much as the choice of model itself. This subsection quantifies that dependency directly. Using the training dataset restricted to Variant A features, each of the six models was evaluated with default hyperparameters under a single random 80:20 split into training and test sets (Appendix Table~\ref{tab:single_random_split_full}) and under a chronological 80:20 split, with the earliest 80\% of records used for training and the remaining 20\% for testing (Appendix Table~\ref{tab:chronological_split_full}). To assess whether the random-split results were sensitive to the particular partition drawn, this procedure was repeated across ten independently reseeded random 80:20 splits, with results reported in (Appendix Table~\ref{tab:repeated_random_split_full}.) Figure~\ref{fig:train_test_random_vs_chronological} summarises training and test performance across all three conditions for each model.

Under the single random split, the four linear models attained test
R\textsuperscript{2} between 0.71 and 0.80, with RMSE between 77.24 and 92.76~L~h\textsuperscript{-1} and PA15 between 60.46\% and 74.37\%, whereas Random Forest and XGBoost attained test R\textsuperscript{2} of 0.99 and 0.98, RMSE of 17.38 and 23.11~L~h\textsuperscript{-1}, and PA15 above 99\% in both cases. Repeating the random split across ten independently reseeded partitions yielded a test RMSE standard deviation below 0.08~L~h\textsuperscript{-1} for every model, with the standard deviation of test R\textsuperscript{2} close to zero across all repetitions, so this result reflects a consistent property of random partitioning applied to this dataset rather than the outcome of a single favourable draw.

Under chronological partitioning, the relative performance of the two model families was inverted. Test R\textsuperscript{2} for Random Forest and XGBoost fell to $-0.36$ and $-0.10$, respectively --- both below the performance of a constant-mean predictor --- with test RMSE rising to 187.81 and 168.98~L~h\textsuperscript{-1} and PA15 falling to 58.18\% and 59.10\%, despite training R\textsuperscript{2} remaining above 0.98 for both models under either partitioning scheme. The linear models showed the opposite trend: test R\textsuperscript{2} for Ridge increased from 0.80 to 0.84, RMSE decreased from 77.24 to 65.02~L~h\textsuperscript{-1}, and PA15 increased from 74.37\% to 82.42\%.

This reversal shows that random and time-aware partitioning answer different questions. A random split places the temporal neighbours of nearly every test observation into the training set, so the resulting score reflects interpolation within already-observed conditions rather than generalisation to unseen ones. A chronological split removes this adjacency, requiring the model to predict a period it has not seen during training. Training and test performance remained close under random partitioning for every model, but diverged sharply for the tree-based ensembles under chronological partitioning: agreement between training and test scores under random splitting is not sufficient evidence of generalisability.

\subsection{Cross-Validation Performance and Hyperparameter Selection}
\begin{figure}[p]
    \centering
    \includegraphics[width=\textwidth, height=0.85\textheight, keepaspectratio]{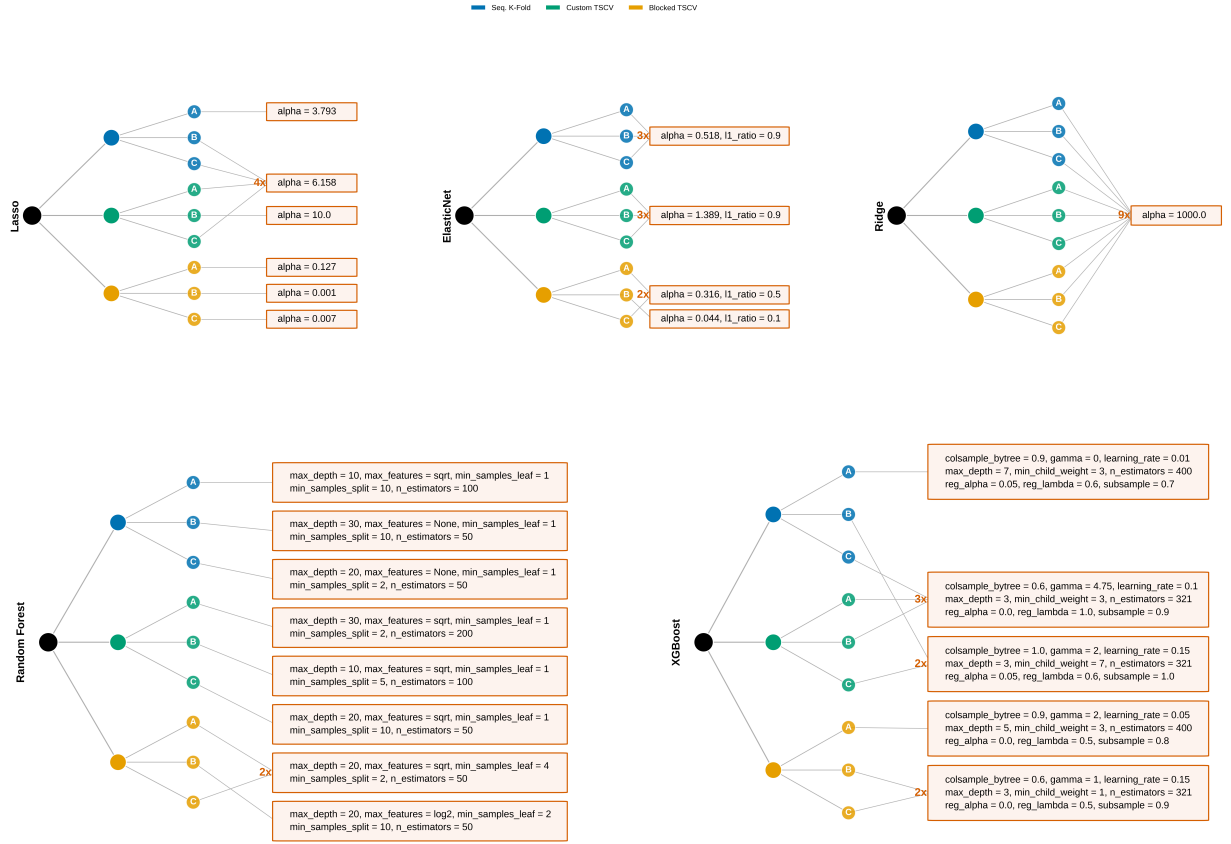}
    \caption{Selected hyperparameters for each model under each
    cross-validation scheme and feature variant. Labels of the form
    $n\times$ mark identical configurations selected in multiple
    combinations.}
    \label{fig:hyperparameter_tree}
\end{figure}

\begin{figure}[htbp]
    \centering
    \includegraphics[width=\textwidth]{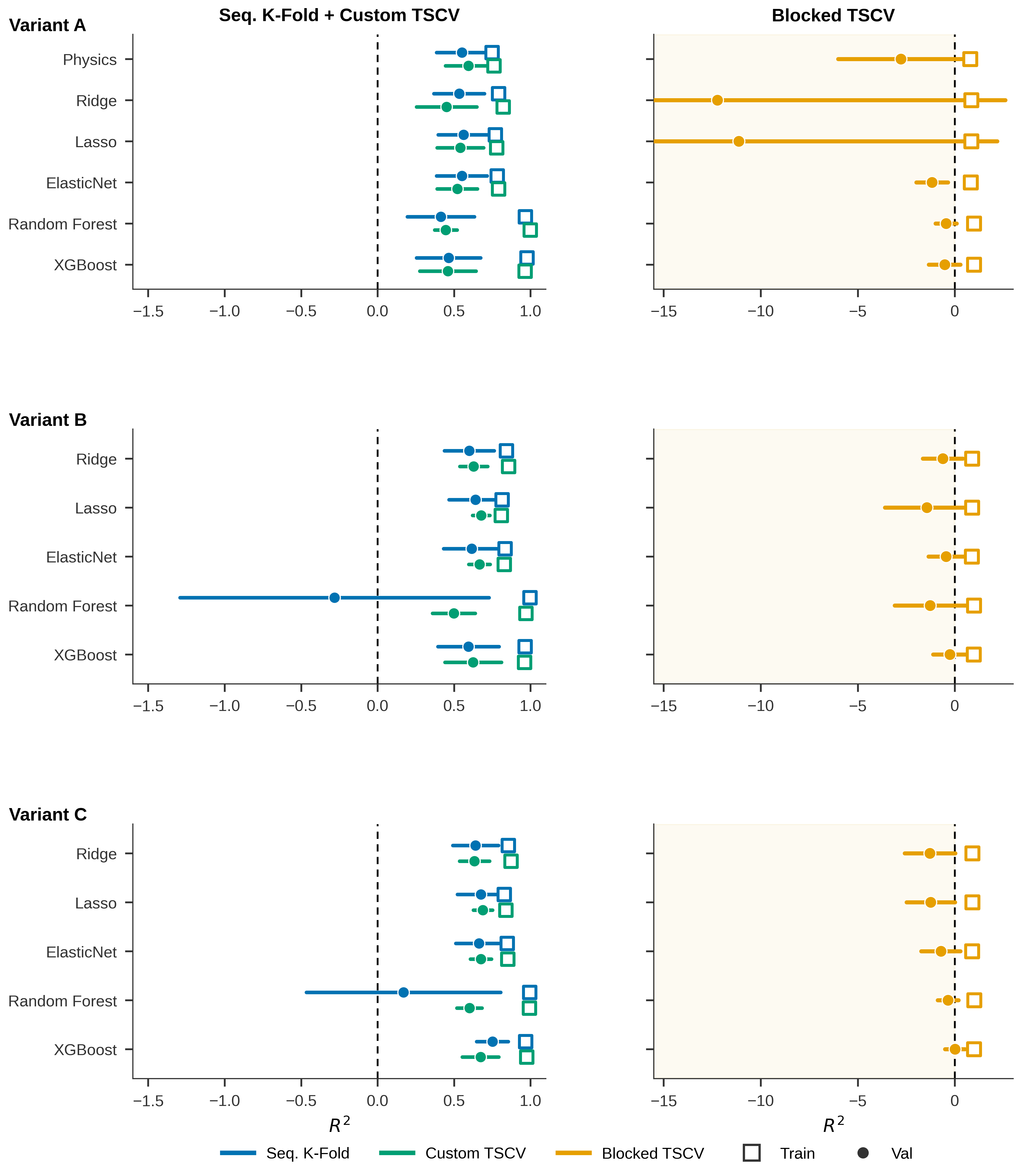}
    \caption{Validation $R^2$ across folds for each model, cross-validation scheme and feature variant. Markers show the mean; whiskers show the fold-to-fold range.}
    \label{fig:cv_stability}
\end{figure}

Having established the need for time-aware evaluation, this subsection
reports performance under the three time-aware cross-validation schemes
described in Section~4.4. Each of the five tunable models was tuned
separately under every combination of scheme and feature variant, giving 45
independent tuning runs; the physics baseline was refit on the training
folds of each scheme for Variant~A. Full training and validation performance
for every combination is reported in Appendix
Table~\ref{tab:cv_performance_all_variants}, with the selected
hyperparameter configurations shown in Figure~\ref{fig:hyperparameter_tree}
and the fold-to-fold spread in validation $R^2$ shown in
Figure~\ref{fig:cv_stability}.

As shown in Figure~\ref{fig:hyperparameter_tree}, Ridge selected the largest
available penalty, $\alpha = 1000$, in all nine runs, so the fitted model
was identical under every scheme and variant. Lasso and ElasticNet selected
moderate penalties under Sequential K-Fold and Custom TSCV (Lasso $\alpha$
between 3.79 and 10.00) and weaker penalties under Blocked TSCV (Lasso
$\alpha$ as low as 0.001). Random Forest was selected close to unconstrained
under all three schemes. XGBoost varied the most, with maximum depth
ranging from 3 to 7 and the number of trees from 321 to 400.

The linear models fitted the training folds moderately well and validated
at a comparable, if lower, level: for Variant~A under Sequential K-Fold,
Ridge reached a training $R^2$ of 0.79 against a validation $R^2$ of 0.53.
The tree-based ensembles fitted the training folds almost exactly without a
corresponding validation result: for Variant~A under Custom TSCV, Random
Forest reached a training $R^2$ above 0.99 and a training RMSE of
9.68~L~h\textsuperscript{-1}, against a validation RMSE of
113.65~L~h\textsuperscript{-1}.

Sequential K-Fold and Custom TSCV produced validation $R^2$ between 0.45 and
0.69 for the linear models across all variants. Blocked TSCV returned a
negative mean validation $R^2$ for every model in every variant except
XGBoost under Variant~C (0.01), with values as large in magnitude as
$-12.23 \pm 14.83$ for Ridge under Variant~A. The physics baseline returned
$-2.78 \pm 3.23$ under the same scheme. As shown in
Figure~\ref{fig:cv_stability}, the fold-to-fold spread in validation $R^2$
was markedly wider under Blocked TSCV than under either Sequential K-Fold
or Custom TSCV for every model.

Validation RMSE under Blocked TSCV was frequently the lowest of the three
schemes despite its negative $R^2$: XGBoost under Variant~C returned a
validation RMSE of 59.10~L~h\textsuperscript{-1} under Blocked TSCV, against
71.94 and 85.82~L~h\textsuperscript{-1} under Sequential K-Fold and Custom
TSCV.


\subsection{Feature Variant Comparison}

\begin{figure}[htbp]
    \centering
    \includegraphics[width=\textwidth]{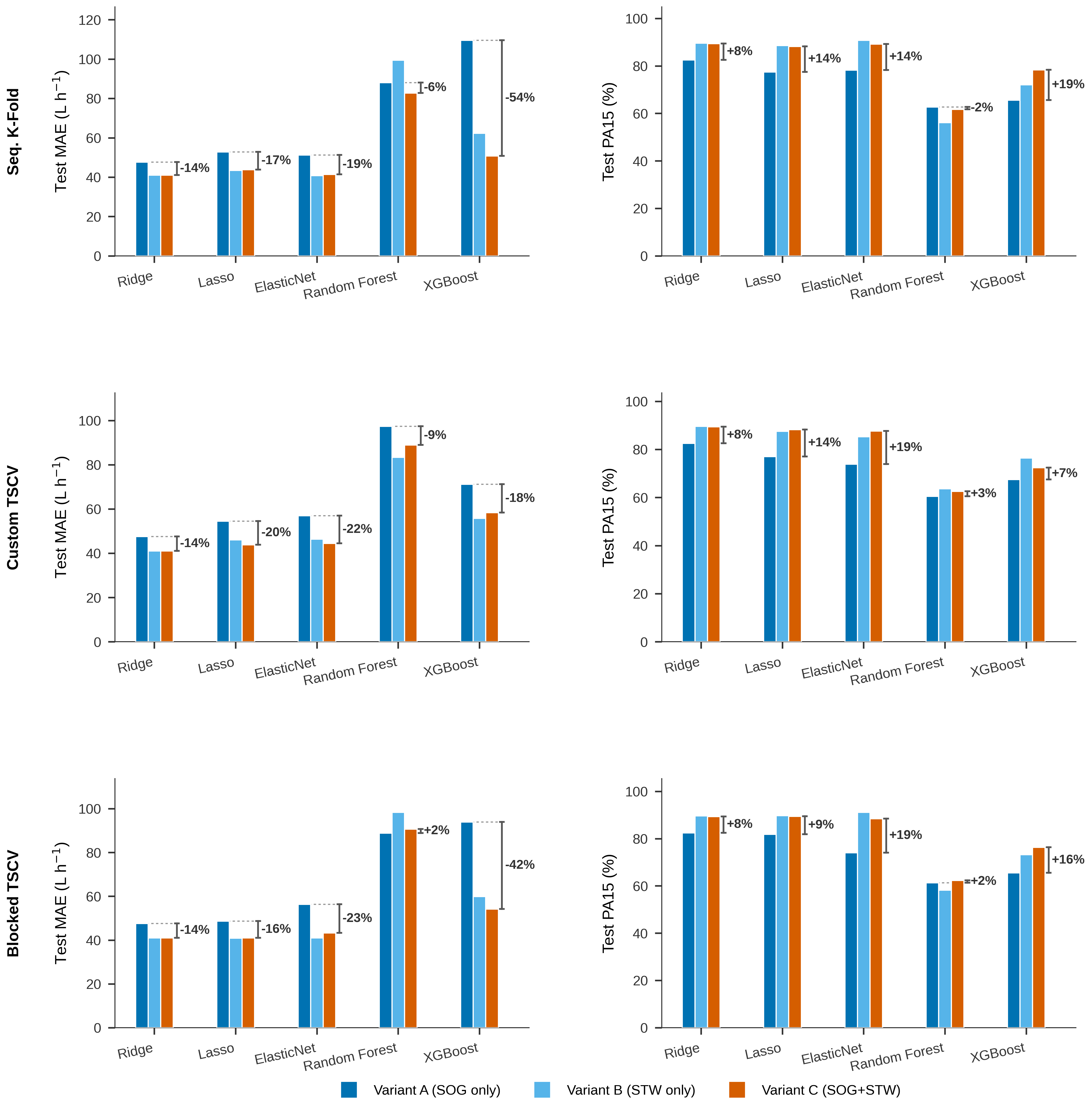}
    \caption{Test MAE and test PA15 by feature variant for each model under
    Sequential K-Fold, Custom TSCV and Blocked TSCV. Percentage changes from
    Variant~A to Variant~C are annotated above each model group.}
    \label{fig:variant_mae_pa15}
\end{figure}

Sections~5.1 and~5.2 held the feature set fixed while varying the validation protocol; this subsection instead holds the protocol fixed and varies the feature set, isolating the contribution of the two speed channels described in Section~4.1. Test MAE and test PA15 were compared across Variants~A, B and C for each model under each cross-validation scheme, with results summarised in Figure~\ref{fig:variant_mae_pa15} and percentage changes from Variant~A to Variant~C annotated above each model group.

Substituting speed over ground with speed through water produced a consistent reduction in test MAE across all three schemes. Under Sequential K-Fold, MAE for the linear models fell by 14--19\% from Variant~A to Variant~C, with corresponding PA15 gains of 8--14 percentage points; reductions of comparable magnitude appeared under Custom TSCV and Blocked TSCV. Since the three variants differ only in the speed signal provided, with all environmental, directional and vessel-state predictors held constant, this improvement is attributable to the speed channel rather than to any other predictor. For the linear models, MAE and PA15 under Variant~B and Variant~C were closely comparable across all three schemes: most of the Variant~A-to-C improvement came from introducing speed through water, with limited additional benefit from also including speed over ground. The tree-based ensembles followed a similar pattern, with one exception: under Sequential K-Fold and Blocked TSCV, XGBoost's Variant~A-to-C reduction in MAE exceeded that of every other model.

\subsection{Hold-Out Test Set Evaluation}
\begin{figure}[htbp]
    \centering
    \includegraphics[width=\textwidth]{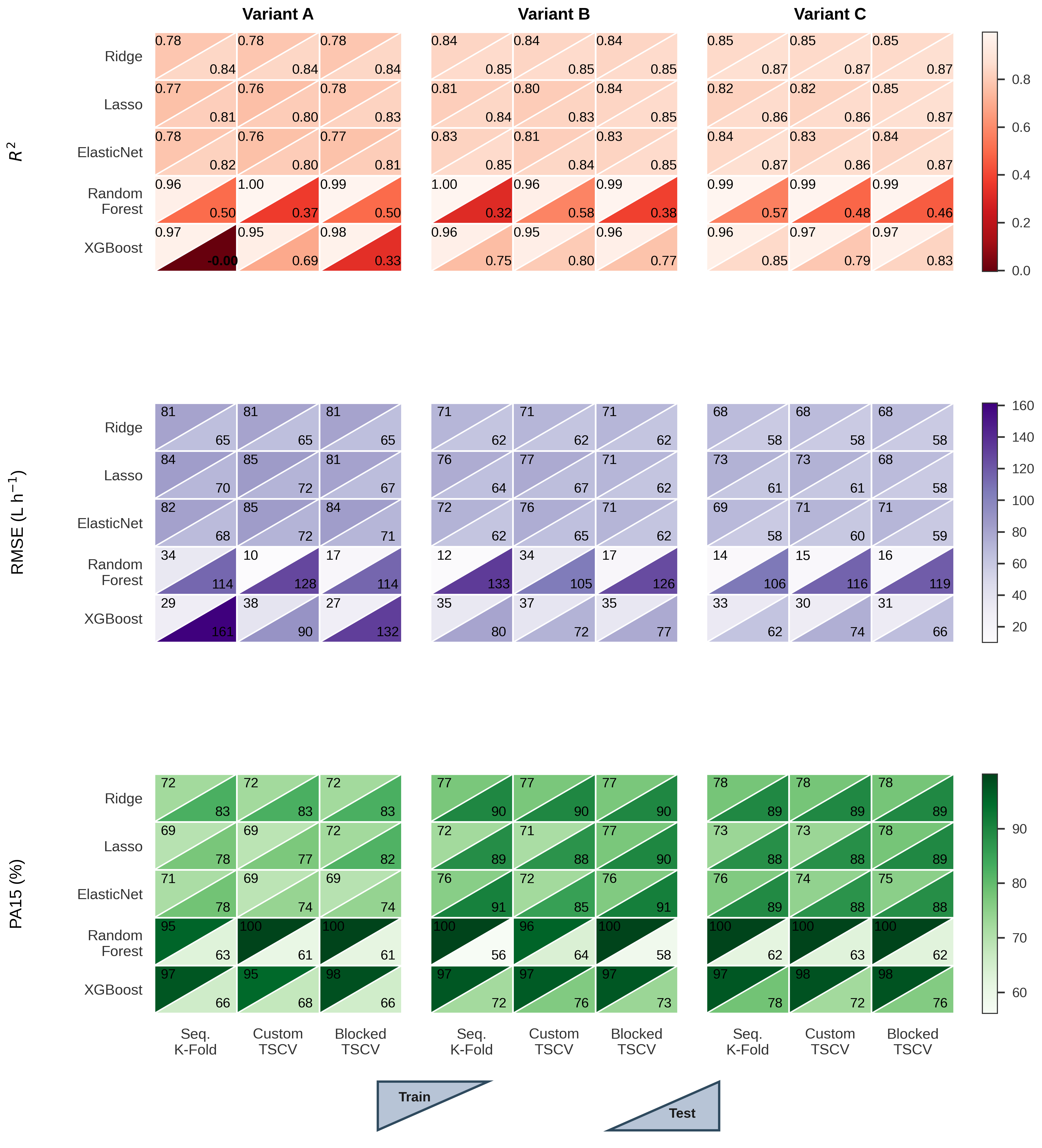}
    \caption{Training and test performance on the chronological hold-out set across all model, cross-validation scheme, and feature variant combinations: R\textsuperscript{2} (top), RMSE~(L~h\textsuperscript{-1}) (middle), and PA15~(\%) (bottom).}
    \label{fig:holdout_heatmap}
\end{figure}

Every tuned configuration from Section~5.2 --- five models under each of three cross-validation schemes, for each of three feature variants, together with MLR and the physics baseline, neither of which depends on a cross-validation scheme --- was refit on the full training partition using its selected hyperparameters and evaluated once on the common chronological hold-out set. Figure~\ref{fig:holdout_heatmap} reports training and test R\textsuperscript{2}, RMSE and PA15 for all resulting configurations, with the full metric set given in Appendix Table~\ref{tab:holdout_full_all_variants}.

Across all three feature variants, the linear models consistently attained the highest test R\textsuperscript{2} and PA15 alongside the lowest test RMSE, while the tree-based ensembles attained the opposite extreme on all three metrics in every variant. The linear models showed a comparatively small gap between training and test values for all three metrics, whereas the tree-based ensembles attained training R\textsuperscript{2} above 0.99 in almost every configuration alongside test R\textsuperscript{2} well below the values attained by the linear models on the same hold-out set.

The dependence of hold-out performance on the cross-validation scheme used for hyperparameter selection is greatest for XGBoost under Variant~A: test R\textsuperscript{2} ranged from 0.688 under Custom TSCV to $-0.004$ under Sequential K-Fold and 0.330 under Blocked TSCV, with test RMSE ranging correspondingly from 89.97 to 161.39~L~h\textsuperscript{-1} and test PA15 from 65.52\% to 67.51\%. Since the feature variant, training partition and evaluation set were identical across these three configurations, this variation is attributable solely to the cross-validation scheme by which the hyperparameters were selected.

\subsection{Statistical Significance of Model Differences}
Figure~\ref{fig:holdout_heatmap} showed that Ridge outperformed both tree-based ensembles across nearly every feature variant and
cross-validation scheme, while Random Forest and XGBoost performed
comparably to one another. To determine whether these differences reflect a systematic effect rather than a feature of a single hold-out period, three pairwise comparisons were tested for statistical significance: Ridge versus XGBoost, Ridge versus Random Forest, and XGBoost versus Random Forest, the last included to establish whether either ensemble held a consistent advantage over the other. Each comparison was evaluated separately for each of the three feature variants and each of the three cross-validation schemes, giving 27 tests in total. For each test, the hold-out set was divided into 100 contiguous, equally sized temporal blocks, and the squared-error differential between the two models was computed within each block. The resulting 100 block-mean differentials were assessed using a Wilcoxon signed-rank test, with a Benjamini--Hochberg false discovery rate correction applied across all 27 tests at $\alpha = 0.05$. Results are reported in Table~\ref{tab:significance_wilcoxon}.

Ridge attained the lower mean squared error in all nine comparisons against XGBoost and all nine against Random Forest, and every one of these eighteen comparisons remained significant after correction. Against Random Forest, Ridge held the lower error in 65 to 77 of the 100 blocks, with mean squared-error reductions ranging from 7182.6 to 13\,931.5 (L~h\textsuperscript{-1})\textsuperscript{2}. Against XGBoost, the advantage was smaller in magnitude but similarly consistent, with Ridge favoured in 62 to 71 blocks and corrected $p$-values ranging from below 0.0001 to 0.0058.

XGBoost attained the lower mean squared error than Random Forest in eight of the nine comparisons, seven of which remained significant after correction. The exception occurred under Variant~A and Sequential K-Fold, where Random Forest held the lower error in 56 of 100 blocks, with a mean squared-error reduction of 1250.0 (L~h\textsuperscript{-1})\textsuperscript{2}; this was the only comparison in the batch that did not reach significance after correction ($p = 0.1799$).

\begin{landscape}
\begin{table}[htbp]
\centering
\scriptsize
\caption{Pairwise statistical comparison of model errors on the chronological hold-out set, using a Wilcoxon signed-rank test on 100 block-mean squared-error differentials, with Benjamini--Hochberg correction applied across all 27 tests ($\alpha = 0.05$).}
\label{tab:significance_wilcoxon}
\begin{threeparttable}
\resizebox{\linewidth}{!}{%
\begin{tabular}{l l l l S[table-format=5.1] S[table-format=2.0] S[table-format=1.4]}
\toprule
{Model Pair} & {Feature Variant} & {Cross-Validation Scheme} & {Model with Lower Error} & {Mean Squared-Error Reduction} & {Blocks with Lower Error (of 100)} & {Wilcoxon Signed-Rank $p$-value} \\
\midrule
\multirow{9}{*}{Ridge vs XGBoost} & A & Sequential K-Fold & Ridge & 9921.0 & 71 & {\text{<}0.0001} \\
 & A & Custom TSCV & Ridge & 4615.7 & 69 & 0.0002 \\
 & A & Blocked TSCV & Ridge & 4897.5 & 68 & {\text{<}0.0001} \\
 & B & Sequential K-Fold & Ridge & 2353.3 & 67 & 0.0002 \\
 & B & Custom TSCV & Ridge & 1686.1 & 68 & 0.0012 \\
 & B & Blocked TSCV & Ridge & 1190.7 & 63 & 0.0058 \\
 & C & Sequential K-Fold & Ridge & 2959.3 & 65 & 0.0001 \\
 & C & Custom TSCV & Ridge & 5339.1 & 70 & {\text{<}0.0001} \\
 & C & Blocked TSCV & Ridge & 2215.9 & 62 & 0.0034 \\
\midrule
\multirow{9}{*}{Ridge vs Random Forest} & A & Sequential K-Fold & Ridge & 8671.0 & 72 & {\text{<}0.0001} \\
 & A & Custom TSCV & Ridge & 12131.0 & 77 & {\text{<}0.0001} \\
 & A & Blocked TSCV & Ridge & 8832.3 & 73 & {\text{<}0.0001} \\
 & B & Sequential K-Fold & Ridge & 13931.5 & 77 & {\text{<}0.0001} \\
 & B & Custom TSCV & Ridge & 7182.6 & 74 & {\text{<}0.0001} \\
 & B & Blocked TSCV & Ridge & 12123.3 & 73 & {\text{<}0.0001} \\
 & C & Sequential K-Fold & Ridge & 7871.8 & 74 & {\text{<}0.0001} \\
 & C & Custom TSCV & Ridge & 9999.6 & 68 & {\text{<}0.0001} \\
 & C & Blocked TSCV & Ridge & 10690.7 & 65 & {\text{<}0.0001} \\
\midrule
\multirow{9}{*}{XGBoost vs Random Forest} & A & Sequential K-Fold & Random Forest & 1250.0 & 56 & 0.1799 \\
 & A & Custom TSCV & XGBoost & 7515.3 & 61 & {\text{<}0.0001} \\
 & A & Blocked TSCV & XGBoost & 3934.7 & 56 & 0.0230 \\
 & B & Sequential K-Fold & XGBoost & 11578.2 & 79 & {\text{<}0.0001} \\
 & B & Custom TSCV & XGBoost & 5496.5 & 60 & {\text{<}0.0001} \\
 & B & Blocked TSCV & XGBoost & 10932.6 & 71 & {\text{<}0.0001} \\
 & C & Sequential K-Fold & XGBoost & 4912.6 & 68 & 0.0003 \\
 & C & Custom TSCV & XGBoost & 4660.5 & 58 & 0.0032 \\
 & C & Blocked TSCV & XGBoost & 8474.8 & 60 & {\text{<}0.0001} \\
\bottomrule
\end{tabular}
}
\end{threeparttable}
\end{table}
\end{landscape}

\subsection{Error Diagnostics and Model Interpretability}
The analyses in this subsection use models refit on the Variant~C training partition, so both speed channels are available and their relative contribution to model predictions can be examined directly.

\subsubsection{Prediction Bias and Extrapolation Behaviour}
\begin{figure}[htbp]
    \centering
    \includegraphics[width=\textwidth]{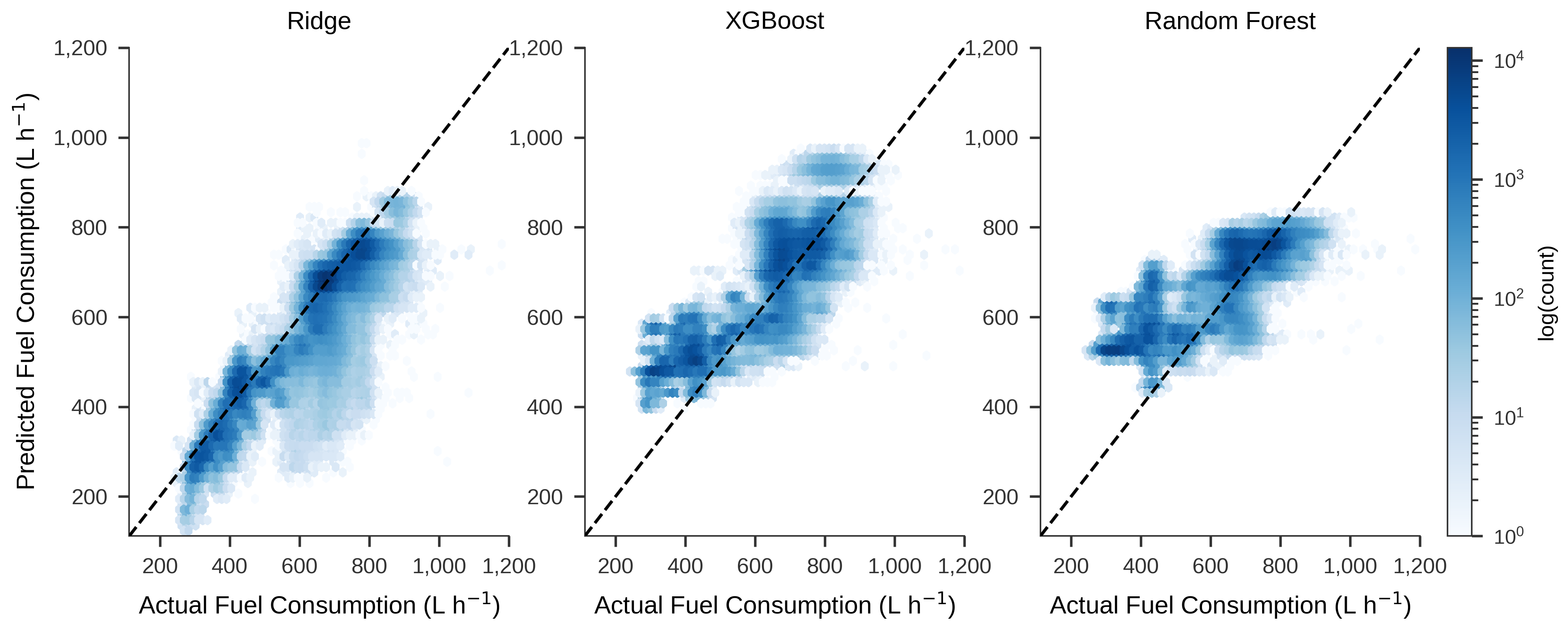}
    \caption{Predicted versus observed fuel consumption for Ridge, XGBoost
    and Random Forest on the Variant~C hold-out set. The dashed line
    indicates perfect agreement.}
    \label{fig:actual_vs_predicted}
\end{figure}

Figure~\ref{fig:actual_vs_predicted} shows predicted against observed fuel consumption for Ridge, XGBoost and Random Forest on the Variant~C hold-out set. Ridge predictions are distributed along the identity line across the full observed range, from below 200 to above 900~L~h\textsuperscript{-1}. XGBoost predictions are concentrated above 400~L~h\textsuperscript{-1}, with few predictions falling below this value despite observed values extending well below it. Random Forest predictions are confined to a narrower band, approximately 450 to 850~L~h\textsuperscript{-1}, and are arranged in distinct horizontal bands rather than distributed continuously along the diagonal.

\begin{figure}[htbp]
    \centering
    \includegraphics[width=\textwidth]{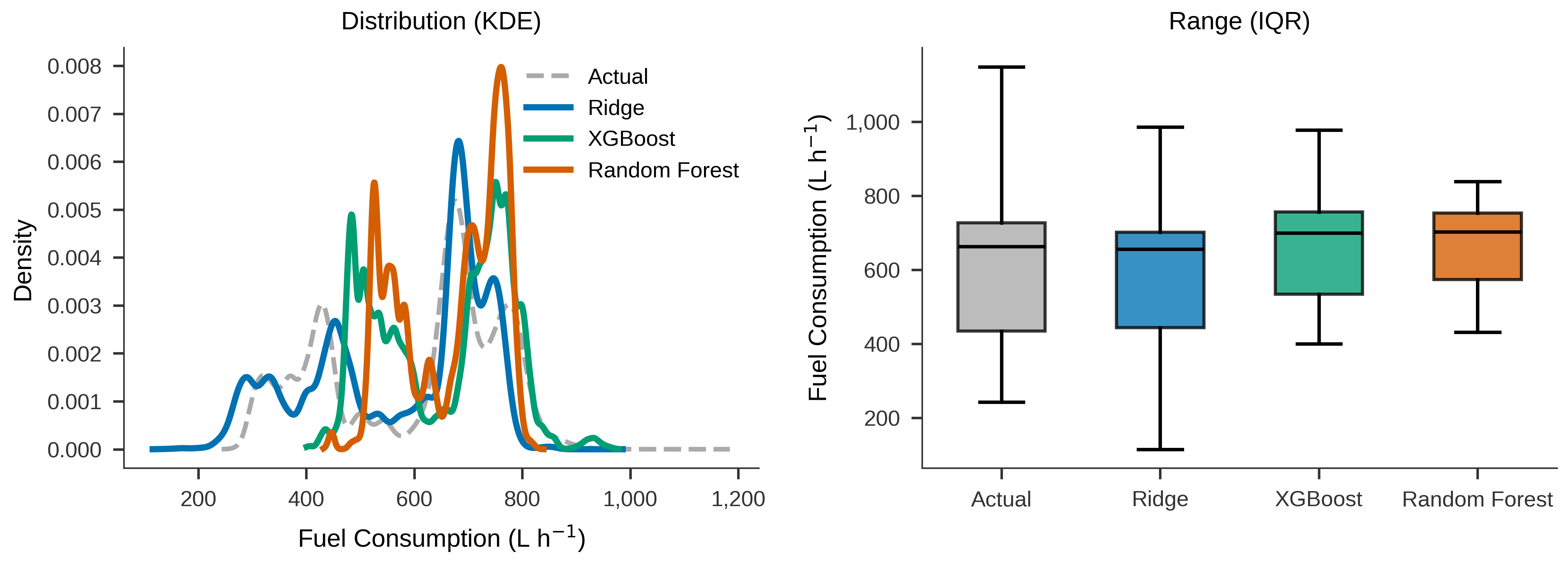}
    \caption{Distribution of predicted fuel consumption for Ridge, XGBoost
    and Random Forest, compared to the observed distribution, on the
    Variant~C hold-out set. Left: kernel density estimate. Right:
    interquartile range and non-outlier range.}
    \label{fig:rf_compression}
\end{figure}

Figure~\ref{fig:rf_compression} compares the distribution of predicted values to the distribution of observed values for the same three models. The observed distribution spans approximately 230 to 1150~L~h\textsuperscript{-1}, with peaks near 430, 620 and
750~L~h\textsuperscript{-1}. Ridge reproduces this distribution closely, including the lower peak. XGBoost and Random Forest place little to no density below approximately 400~L~h\textsuperscript{-1}, and Random Forest shows a pronounced peak near 755~L~h\textsuperscript{-1} that exceeds the density of the observed distribution at the same value. The observed interquartile range (IQR) spans approximately 430 to 730~L~h\textsuperscript{-1}, with the full non-outlier range spanning approximately 240 to 1150~L~h\textsuperscript{-1}; Ridge reproduces both closely, while Random Forest returns a narrower interquartile range of approximately 555 to 770~L~h\textsuperscript{-1} and a non-outlier range spanning only 445 to 845~L~h\textsuperscript{-1}.


\subsubsection{Feature Attribution via Coefficients and SHAP Values}

\begin{figure}[htbp]
    \centering
    \includegraphics[width=\textwidth]{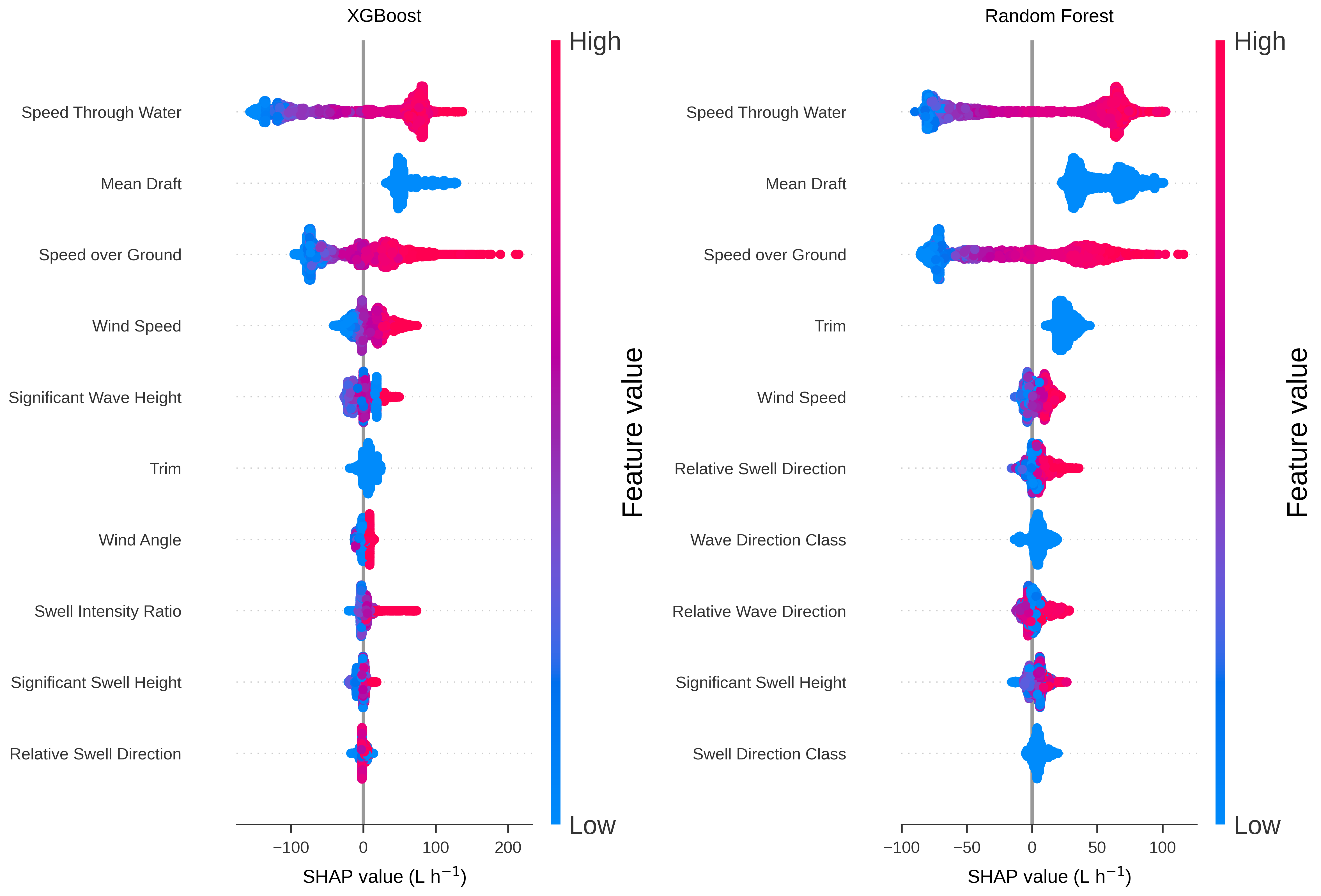}
    \caption{SHAP value distributions for XGBoost and Random Forest on the
    Variant~C hold-out set. Each point represents one observation; colour
    indicates the corresponding feature value.}
    \label{fig:shap_beeswarm}
\end{figure}

Figure~\ref{fig:shap_beeswarm} shows SHAP value distributions for XGBoost and Random Forest. In both models, speed through water, mean draft and speed over ground rank among the three highest-magnitude features. Wind speed and trim follow among the next-highest-ranked features for both models. For XGBoost, significant wave height, wind angle, swell intensity ratio, significant swell height and relative swell direction complete the ten highest-ranked features. For Random Forest, relative swell direction, wave direction class, relative wave direction, significant swell height and swell direction class complete the ten highest-ranked features. Mean draft contributions are uniformly positive in both models, spanning approximately 20 to 130~L~h\textsuperscript{-1} for XGBoost and 20 to 105~L~h\textsuperscript{-1} for Random Forest.

\begin{figure}[htbp]
    \centering
    \begin{minipage}{0.48\textwidth}
        \centering
        \includegraphics[width=\textwidth]{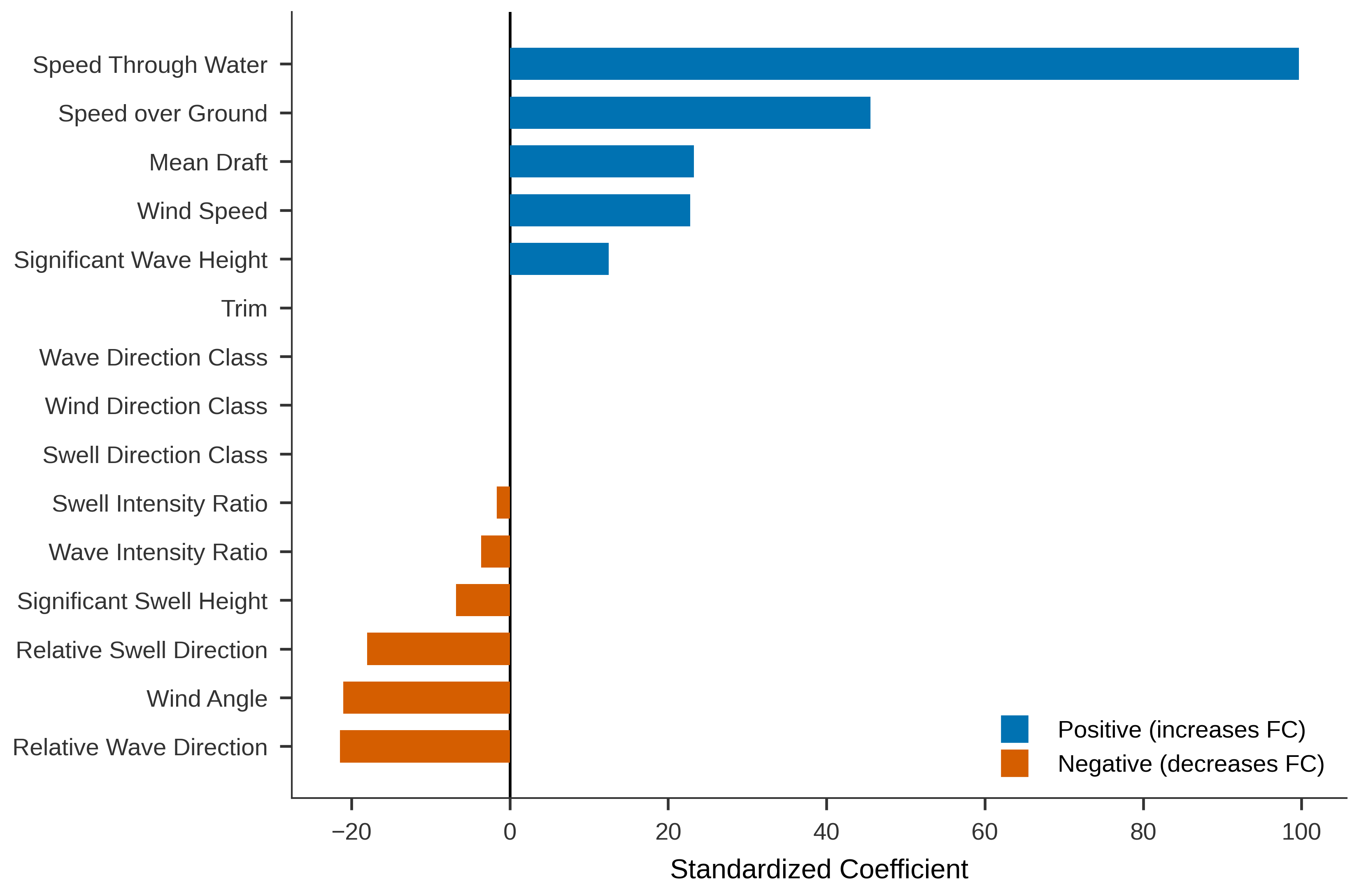}
    \end{minipage}
    \hfill
    \begin{minipage}{0.48\textwidth}
        \centering
        \includegraphics[width=\textwidth]{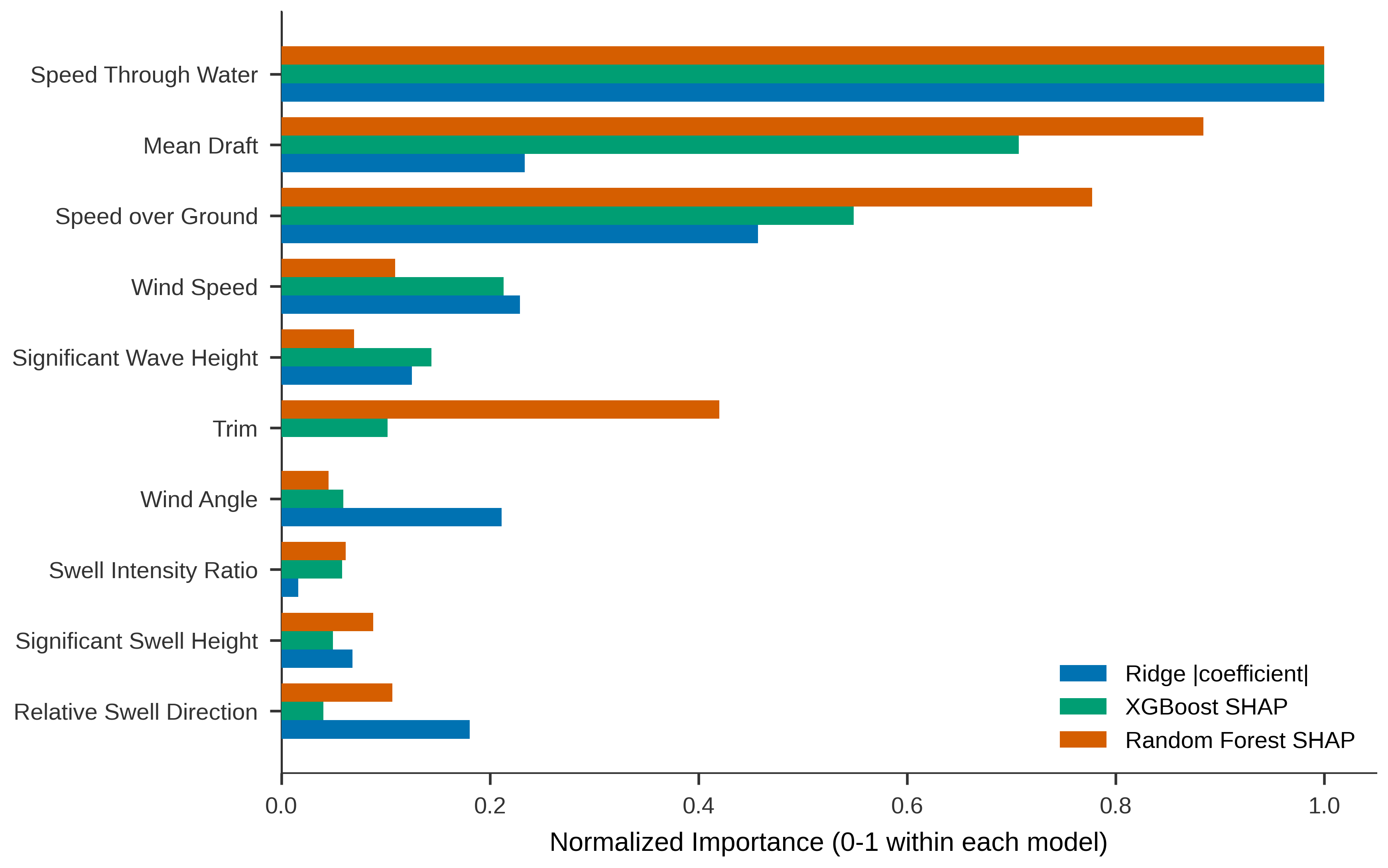}
    \end{minipage}
    \caption{Left: standardized Ridge coefficients for the Variant~C model,
    signed by direction of effect on predicted fuel consumption. Right:
    normalized feature importance (0--1 within each model) for Ridge,
    XGBoost and Random Forest.}
    \label{fig:ridge_and_importance}
\end{figure}

Figure~\ref{fig:ridge_and_importance} reports the standardized Ridge coefficients alongside a normalized comparison of feature importance across all three models. Speed through water carries the largest Ridge coefficient, followed by speed over ground, mean draft and wind speed, all with positive sign; significant wave height carries a smaller positive coefficient, and trim carries a coefficient close to zero. Relative wave direction, wind angle and relative swell direction carry the largest negative coefficients, followed by significant swell height, wave intensity ratio and swell intensity ratio. In the normalized comparison, speed through water ranks highest for all three models, and mean draft and speed over ground rank among the top three for XGBoost and Random Forest; trim ranks substantially higher for the tree-based ensembles than for Ridge, while wind angle and relative swell direction rank higher for Ridge than for either ensemble.

\section{Discussion}

\subsection{Model Flexibility and Robustness to Validation Design}

The penalised linear models showed limited dependence on the cross-validation scheme used for hyperparameter selection. Ridge selected the same regularisation strength ($\alpha = 1000$, the upper bound of the search grid) under every scheme and feature variant, yielding an identical fitted model in all nine cases; at this sample size, a penalty of this magnitude leaves the fitted coefficients effectively indistinguishable from those of unpenalised linear regression (MLR), so Ridge's stability across schemes reflects the absence of an active penalty rather than regularisation conferring robustness. The hold-out RMSE of Lasso and ElasticNet varied by no more than 4.77~L~h$^{-1}$ across schemes within any feature variant. The tree-based ensembles showed substantially greater dependence on this choice: hold-out RMSE for Random Forest varied by 12.55 to 28.33~L~h$^{-1}$ according to the scheme by which its hyperparameters were selected, and for XGBoost by 8.09 to 71.42~L~h$^{-1}$, with the largest difference under Variant A, where hold-out RMSE ranged from 89.97~L~h$^{-1}$ under Custom TSCV to 161.39~L~h$^{-1}$ under Sequential K-Fold despite an identical feature set, training partition, and evaluation set across the three configurations. This contrast reflects a difference in model flexibility rather than a fixed property of any individual scheme: the ensembles have enough capacity to fit structure specific to whichever subset of observations made up each scheme's validation folds, so a change in fold composition yields a materially different tuned configuration, whereas the linear models lack the capacity to fit such fold-specific structure and were correspondingly insensitive to it.

Two properties of the ensembles account for this sensitivity. Tree predictions are bounded by the training response range, so hold-out conditions outside this range are predicted at the nearest bound rather than extrapolated, unlike a linear model's unbounded coefficients. Mean draft is also constant at 5.995~m throughout the chronological hold-out set, despite spanning several discrete ballast configurations in training — a loading condition changes only between port calls, and so can readily remain fixed within a single chronological partition. Since draft ranks among the three highest-magnitude SHAP features for both ensembles (Section~5.6.2), the draft-based splits learned in training carry no discriminating power over this hold-out period, leaving hold-out performance to depend on how heavily each tuned configuration relied on draft and on the training operating range more broadly.

\subsection{Evaluation Metrics for Decision Support: Beyond $R^2$}
Because model tuning throughout this study was scored on validation RMSE rather than $R^2$ (Section~4.6), the negative validation $R^2$ observed under Blocked TSCV (Section~5.2) does not indicate that the affected models failed to satisfy the criterion by which they were selected. $R^2$ expresses the proportion of variance in the response explained by the model, and is therefore defined relative to the variance of the response within the fold being scored, rather than to any absolute error scale.
This dependence makes it particularly sensitive to fold construction under Blocked TSCV, where each block is trained and validated entirely within itself on a short, contiguous segment of the record (Section~4.4.4); such segments span a narrower range of operating conditions, and consequently show lower response variance, than the larger cumulative windows constructed by Sequential K-Fold and Custom TSCV. Where response variance within a fold is small, even a modest absolute error is disproportionately amplified in the resulting $R^2$, so the statistic becomes unstable irrespective of the accuracy of the underlying predictions. The physics baseline, with only three fitted parameters and no capacity to overfit, returned a comparably negative validation $R^2$ under Blocked TSCV ($-2.78 \pm 3.23$) to the tuned ensembles — confirming that this instability reflects the validation scheme, not model behaviour.

This behaviour illustrates a broader limitation of variance-relative statistics for reporting model performance in a decision-support context. PA15 and Mean Accuracy (Section~4.5) are computed from the absolute deviation between prediction and observation, without reference to the variance of the evaluation window, and are therefore not subject to the instability described above: a model's PA15 reflects how often its predictions fell within an operationally meaningful margin, irrespective of how much the underlying conditions happened to vary over the period being scored. $R^2$, by contrast, answers a different question --- how much of the variance within that specific window was explained --- which bears an indirect relationship, at best, to whether a given prediction can be trusted operationally. For a decision support system intended to inform voyage-planning choices, bounded, absolute-error metrics of this kind give a more stable and directly interpretable basis for communicating model reliability to an operational user than variance-relative statistics such as $R^2$.

This distinction is evident in the feature-variant comparison. Variant A, the only configuration available at the voyage-planning stage, reached a test PA15 of approximately 83\,\% under Ridge, compared with approximately 89--90\,\% for Variants B and C, which additionally incorporate speed through water. A single $R^2$ value for Variant A would not, by itself, convey the operational significance of this gap: PA15 states directly that roughly one in six predictions under Variant A falls outside an acceptable margin, a figure a voyage planner can weigh against their own risk tolerance in a manner that a variance-explained statistic cannot readily supply.

\subsection{Limitations}
Time-aware validation, as applied throughout this study, substantially reduces the temporal leakage demonstrated in Section~5.1, but does not necessarily eliminate it. Chronological splitting guarantees that no test observation precedes any observation used to train the corresponding model; it does not guarantee that the operating conditions on either side of a partition boundary differ meaningfully from one another. Because steady-state segments retained after preprocessing (Section~3.2) were frequently adjacent in real time prior to filtering, the sample immediately preceding a given fold boundary and the sample immediately following it may still describe a near-identical vessel state. The two protocols differ, however, in the extent of the record over which this condition obtains. Under chronological partitioning it is confined to the immediate neighbourhood of the single hold-out boundary, and of the $k$ fold boundaries internal to each cross-validation scheme, leaving the majority of the test partition separated from the training record by intervals of hours to months. Under random partitioning, the temporal neighbours of nearly every test observation are assigned to the training set, so the condition obtains across the whole of the test partition rather than at a small number of isolated boundaries. Residual adjacency of the former kind is difficult to exclude in any validation scheme applied to continuous operational data, and constitutes a limitation of the evaluation rather than a defect specific to the schemes examined here; it is not, however, of a magnitude capable of
accounting for the differences reported in Section~5.1.

The dataset spans a single vessel over a single year (August 2024 to June 2025), which confines both the training data and the chronological hold-out period to whatever range of conditions occurred within that interval. This scope is consistent with the constant mean draft observed throughout the hold-out set (Section~6.1): a hold-out period drawn from a single year may be short enough to fall within one loading condition rather than spanning several. The same limited scope compounds the coarser temporal and spatial resolution of the ERA5 hindcast fields (hourly, 0.5\textdegree) relative to the 1~Hz sensor record, restricting the variability available to the environmental predictors and offering a plausible explanation for their comparatively modest SHAP attribution (Section~5.6.2). Extending the dataset across additional vessels, seasons, and loading conditions would test whether the reported performance figures hold beyond this single case, without bearing on the validity of
the comparative validation methodology itself.

\section{Conclusions}
This study evaluated six regression models and a physics-based baseline on approximately 3.89 million steady-state 1 Hz records from the CCGS \textit{Sir Wilfrid Laurier}, collected between August 2024 and June 2025. Three feature variants and three time-aware cross-validation schemes were tested, and every resulting model was evaluated on the same chronological hold-out set, so the comparisons reported below isolate the effect of model choice and validation protocol.

Random partitioning of the training dataset produced substantially higher apparent test performance than chronological partitioning. Under a single random 80:20 split, Random Forest and XGBoost attained test $R^2$ of 0.99 and 0.98, with RMSE of 17.38 and 23.11~L~h$^{-1}$. Under chronological partitioning of the same training dataset, the two models attained test $R^2$ of $-0.36$ and $-0.10$, with RMSE rising to 187.81 and 168.98~L~h$^{-1}$. Training $R^2$ remained above 0.98 for both models under both partitioning schemes, so both models learned their training data equally well in each case: the difference lies entirely in test performance, and reflects a failure to generalise rather than a failure to learn. Ten independently reseeded random splits gave a test RMSE standard deviation below 0.08~L~h$^{-1}$, so this result was consistent rather than dependent on any single split. At this sampling frequency, a high test score obtained under random partitioning does not indicate that a model will generalise to unseen conditions: because adjacent 1 Hz observations are nearly identical, random splitting places near-duplicates of most test observations into the training set, so the model has effectively already seen the conditions it is being tested on.

Under time-aware validation, Ridge attained the lower mean squared error than both Random Forest and XGBoost in all eighteen comparisons across every feature variant and cross-validation scheme, and all eighteen differences were statistically significant under a Wilcoxon signed-rank test on block-mean squared-error differentials, Benjamini--Hochberg corrected at $\alpha = 0.05$. Ridge selected the strongest available penalty in every run, making its fit effectively equivalent to unpenalised linear regression; the advantage therefore reflects the suitability of a linear model to this data, not a benefit from regularisation.

The physics-based baseline, fitted with only three coefficients, attained a test $R^2$ of 0.71 and RMSE of 86.09~L~h$^{-1}$ under Variant A, outperforming the tuned tree-based ensembles but remaining below the linear models under the same conditions.

The weaker hold-out performance of Random Forest and XGBoost likely reflects the limited operating range of this single-vessel, single-year dataset rather than a general limitation of tree-based methods, since both ensembles are bounded by the range of values seen during training. A dataset spanning more vessels, seasons, and loading conditions may narrow or reverse this gap.

The feature-variant comparison quantified the accuracy cost of restricting inputs to those available at the voyage-planning stage. Variant A, the only configuration available before departure, attained a Ridge test PA15 of approximately 83\%, compared with 89--90\% for Variants B and C, which additionally incorporate speed through water. This gap is the accuracy a voyage planner forgoes in exchange for a prediction that can be made before the leg is sailed.

\section{Future Work}
The present study is confined to a single vessel over an eleven-month period, which limits the range of operating and loading conditions available for both training and evaluation. Extending the analysis to additional vessels, seasons, and loading conditions would test whether the bounded extrapolation behaviour and correspondingly weaker hold-out performance observed for the tree-based ensembles in Section~6 are a consequence of this restricted operating envelope or a more general property of the model class.

Neural network architectures were not evaluated in this study and represent a natural extension: recurrent and temporal-convolutional networks can exploit the sequential structure of high-frequency operational data directly, rather than treating each observation as independent, while physics-informed neural networks that embed the resistance relations reviewed in Section~2.1.1 as constraints on the learned function may combine the extrapolation behaviour of the physics-based baseline with the flexibility of a data-driven model. Given the sensitivity of flexible model classes to validation protocol demonstrated in this study, any accuracy gains reported for these architectures should be confirmed under chronological rather than random partitioning before being taken as evidence of genuine improvement.

Finally, the present analysis predicts fuel consumption at each 1 Hz observation independently; formulating the prediction task at the level of discrete voyage segments would align the model output more closely with the resolution at which voyage-planning decisions are actually made.

\section*{Data Availability}
The data that support the findings of this study were obtained from the  Canadian Coast Guard and the National Research Council Canada under  research agreements containing confidentiality provisions. Owing to the  sensitive and proprietary nature of these operational datasets, they are  not publicly available. Derived and aggregated results supporting the  conclusions of this paper are available from the corresponding author  upon reasonable request, subject to institutional approval. Code used  for data processing, model development, and analysis is available at  \url{https://github.com/Samar-Maris/ship-fuel-consumption-ml}.

\section*{CRediT authorship contribution statement}

\textbf{Samarasimha Reddy Chittamuru:} Conceptualization, Methodology, Data curation, Formal analysis, Software, Validation, Visualization, Investigation, Writing – original draft.  
\textbf{Ayhan Akinturk:} Supervision, Methodology, Validation, Writing – review and editing.  
\textbf{Allison Kennedy:} Project administration, Funding acquisition, Resources, Writing – review and editing.  
\textbf{Joshua Barnes:} Conceptualization, Methodology, Writing – review and editing.  
\textbf{Matthew Hamilton:} Supervision, Conceptualization, Methodology, Investigation, Writing – review and editing.

\section*{Funding Sources}
This research was supported by the National Research Council Canada (NRC) through the Greening Government Fund (GGF) project titled “Operational Data Analytics: A Demonstration of Benefits to Greening Government Marine Fleets” and the Office of Energy Research and Development (OERD) project titled “Quantifying Vessel Performance: Towards Green Decision Making” (OERD Project code: NRC-23-138).

\section*{Acknowledgements}

The authors gratefully acknowledge the Canadian Coast Guard (CCG) for their collaboration and continued support throughout this research.
The authors especially thank the officers and crew of the CCGS \textit{Sir Wilfrid Laurier} for their assistance and effort in facilitating data collection and operational coordination.

\bibliographystyle{elsarticle-num}
\bibliography{references}

\clearpage
\appendix
\section{Full-Metric Validation Tables}
\label{sec:appendixA}

\begin{landscape}
\begin{table}[htbp]
\centering
\scriptsize
\caption{Model performance under a single random 80:20 train--test split (Variant A), using default (untuned) hyperparameters.}
\label{tab:single_random_split_full}
\begin{threeparttable}
\resizebox{\linewidth}{!}{%
\begin{tabular}{l S[table-format=1.3] S[table-format=3.2] S[table-format=3.2] S[table-format=2.2] S[table-format=2.2]
                  S[table-format=1.3] S[table-format=3.2] S[table-format=3.2] S[table-format=2.2] S[table-format=2.2]}
\toprule
{Model} & {Train $R^2$} & {Train RMSE} & {Train MAE} & {Train Accuracy (\%)} & {Train PA15 (\%)} & {Test $R^2$} & {Test RMSE} & {Test MAE} & {Test Accuracy (\%)} & {Test PA15 (\%)} \\
\midrule
MLR & 0.800 & 77.12 & 57.23 & 88.90 & 74.40 & 0.799 & 77.24 & 57.35 & 88.87 & 74.37 \\
Ridge & 0.800 & 77.12 & 57.23 & 88.90 & 74.40 & 0.799 & 77.24 & 57.35 & 88.87 & 74.37 \\
Lasso & 0.790 & 78.97 & 59.18 & 88.49 & 72.66 & 0.789 & 79.07 & 59.27 & 88.47 & 72.62 \\
ElasticNet & 0.710 & 92.80 & 76.44 & 85.34 & 60.42 & 0.710 & 92.76 & 76.40 & 85.34 & 60.46 \\
Random Forest & 0.999 & 6.59 & 4.29 & 99.27 & 99.99 & 0.990 & 17.38 & 11.46 & 98.06 & 99.86 \\
XGBoost & 0.982 & 23.01 & 16.10 & 97.13 & 99.36 & 0.982 & 23.11 & 16.14 & 97.12 & 99.35 \\
\bottomrule
\end{tabular}
}
\end{threeparttable}
\end{table}

\begin{table}[htbp]
\centering
\scriptsize
\caption{Model performance across 10 independently reseeded random 80:20 splits (Variant A), using default (untuned) hyperparameters. Values are mean $\pm$ standard deviation across the 10 repetitions. Full metric set.}
\label{tab:repeated_random_split_full}
\begin{threeparttable}
\resizebox{\linewidth}{!}{%
\begin{tabular}{l S[separate-uncertainty, table-format=1.3(1)] S[separate-uncertainty, table-format=2.2(2)] S[separate-uncertainty, table-format=2.2(2)] S[separate-uncertainty, table-format=2.2(1)] S[separate-uncertainty, table-format=2.2(1)]
                  S[separate-uncertainty, table-format=1.3(1)] S[separate-uncertainty, table-format=2.2(2)] S[separate-uncertainty, table-format=2.2(2)] S[separate-uncertainty, table-format=2.2(1)] S[separate-uncertainty, table-format=2.2(1)]}
\toprule
{Model} & {Train $R^2$} & {Train RMSE} & {Train MAE} & {Train Accuracy (\%)} & {Train PA15 (\%)} & {Test $R^2$} & {Test RMSE} & {Test MAE} & {Test Accuracy (\%)} & {Test PA15 (\%)} \\
\midrule
MLR & 0.800(0) & 77.15(2) & 57.26(1) & 88.89(0) & 74.39(1) & 0.800(0) & 77.14(7) & 57.24(3) & 88.90(1) & 74.41(3) \\
Ridge & 0.800(0) & 77.15(2) & 57.26(1) & 88.89(0) & 74.39(1) & 0.800(0) & 77.14(7) & 57.24(3) & 88.90(1) & 74.41(3) \\
Lasso & 0.790(0) & 78.99(2) & 59.21(1) & 88.48(0) & 72.64(1) & 0.790(0) & 78.98(7) & 59.18(3) & 88.49(1) & 72.67(4) \\
ElasticNet & 0.710(0) & 92.80(1) & 76.45(1) & 85.34(0) & 60.42(1) & 0.710(0) & 92.81(7) & 76.43(5) & 85.34(1) & 60.43(4) \\
Random Forest & 0.999(0) & 6.59(0) & 4.29(0) & 99.27(0) & 99.99(0) & 0.990(0) & 17.33(4) & 11.45(1) & 98.06(0) & 99.86(0) \\
XGBoost & 0.982(0) & 23.21(8) & 16.22(5) & 97.10(1) & 99.34(1) & 0.982(0) & 23.26(8) & 16.25(5) & 97.10(1) & 99.33(1) \\
\bottomrule
\end{tabular}
}
\end{threeparttable}
\end{table}

\begin{table}[htbp]
\centering
\scriptsize
\caption{Model performance under a chronological hold-out split (Variant A): the earliest 80\% of the time-ordered data for training, the most recent 20\% for testing, with no reshuffling. Default (untuned) hyperparameters throughout. Full metric set.}
\label{tab:chronological_split_full}
\begin{threeparttable}
\resizebox{\linewidth}{!}{%
\begin{tabular}{l S[table-format=1.3] S[table-format=3.2] S[table-format=3.2] S[table-format=2.2] S[table-format=2.2]
                  S[table-format=-1.3] S[table-format=3.2] S[table-format=3.2] S[table-format=2.2] S[table-format=2.2]}
\toprule
{Model} & {Train $R^2$} & {Train RMSE} & {Train MAE} & {Train Accuracy (\%)} & {Train PA15 (\%)} & {Test $R^2$} & {Test RMSE} & {Test MAE} & {Test Accuracy (\%)} & {Test PA15 (\%)} \\
\midrule
\textit{Physics} & 0.747 & 87.63 & 66.34 & 87.33 & 69.11 & 0.714 & 86.09 & 70.51 & 86.45 & 67.23 \\
\midrule
MLR & 0.785 & 80.81 & 60.33 & 88.21 & 72.40 & 0.837 & 65.02 & 47.66 & 91.58 & 82.49 \\
Ridge & 0.785 & 80.81 & 60.33 & 88.21 & 72.40 & 0.837 & 65.02 & 47.66 & 91.58 & 82.49 \\
Lasso & 0.778 & 82.05 & 61.95 & 87.89 & 71.23 & 0.820 & 68.35 & 50.57 & 90.89 & 79.51 \\
ElasticNet & 0.702 & 95.08 & 77.60 & 85.22 & 59.28 & 0.716 & 85.81 & 72.68 & 85.79 & 65.31 \\
Random Forest & 0.999 & 6.32 & 4.01 & 99.31 & 99.99 & -0.360 & 187.81 & 131.37 & 69.67 & 58.18 \\
XGBoost & 0.984 & 22.08 & 15.30 & 97.23 & 99.46 & -0.101 & 168.98 & 117.33 & 73.11 & 59.10 \\
\bottomrule
\end{tabular}
}
\end{threeparttable}
\end{table}
\end{landscape}

\begin{landscape}
\begin{table}[htbp]
\centering
\tiny
\renewcommand{\arraystretch}{1.15}
\caption{Cross-validation performance across all three feature variants: mean $\pm$ standard deviation across folds, training and validation portions, under each cross-validation scheme. The physics baseline is available only for Variant A.}
\label{tab:cv_performance_all_variants}
\begin{threeparttable}
\begin{tabular}{l l l
    S[separate-uncertainty, table-format=1.3(2)]
    S[separate-uncertainty, table-format=2.2(3)]
    S[separate-uncertainty, table-format=2.2(3)]
    S[separate-uncertainty, table-format=-2.3(5)]
    S[separate-uncertainty, table-format=3.2(4)]
    S[separate-uncertainty, table-format=3.2(4)] }
\toprule
{Variant} & {Model} & {CV Scheme} & {Train $R^2$} & {Train RMSE} & {Train MAE} & {Val $R^2$} & {Val RMSE} & {Val MAE} \\
\midrule
\multirow{18}{*}{A} & \multirow{3}{*}{\textit{Physics}} & Sequential K-Fold & 0.748(16) & 86.70(219) & 65.51(193) & 0.551(166) & 96.18(1046) & 74.26(981) \\
 &  & Custom TSCV & 0.761(8) & 82.96(60) & 62.87(63) & 0.594(149) & 94.14(1153) & 69.94(839) \\
 &  & Blocked TSCV & 0.786(27) & 79.89(665) & 59.48(477) & -2.784(3233) & 107.30(2500) & 91.57(2697) \\
\cmidrule(lr){2-9}
 & \multirow{3}{*}{Ridge} & Sequential K-Fold & 0.790(8) & 79.11(198) & 59.08(131) & 0.533(165) & 98.18(978) & 76.06(785) \\
 &  & Custom TSCV & 0.820(18) & 71.96(456) & 54.36(353) & 0.451(197) & 109.75(1160) & 85.13(1041) \\
 &  & Blocked TSCV & 0.849(16) & 67.00(466) & 48.60(153) & -12.233(14828) & 157.02(5101) & 134.17(4927) \\
\cmidrule(lr){2-9}
 & \multirow{3}{*}{Lasso} & Sequential K-Fold & 0.770(10) & 82.89(167) & 63.01(144) & 0.563(167) & 94.68(1013) & 73.25(904) \\
 &  & Custom TSCV & 0.778(14) & 80.05(351) & 61.93(250) & 0.541(152) & 100.78(1109) & 77.85(910) \\
 &  & Blocked TSCV & 0.848(16) & 67.28(468) & 48.85(153) & -11.131(13320) & 152.74(4615) & 130.05(4401) \\
\cmidrule(lr){2-9}
 & \multirow{3}{*}{ElasticNet} & Sequential K-Fold & 0.782(8) & 80.59(191) & 61.02(122) & 0.551(166) & 96.29(1052) & 75.53(910) \\
 &  & Custom TSCV & 0.790(15) & 77.77(376) & 60.52(300) & 0.521(131) & 103.86(1248) & 81.88(1156) \\
 &  & Blocked TSCV & 0.822(6) & 72.94(292) & 54.73(118) & -1.171(820) & 94.84(2780) & 79.24(3139) \\
\cmidrule(lr){2-9}
 & \multirow{3}{*}{Random Forest} & Sequential K-Fold & 0.965(2) & 32.34(103) & 23.36(105) & 0.414(220) & 110.59(1535) & 88.67(1403) \\
 &  & Custom TSCV & 0.997(0) & 9.68(36) & 6.64(42) & 0.446(74) & 113.65(1728) & 88.00(1833) \\
 &  & Blocked TSCV & 0.991(3) & 15.96(294) & 10.94(222) & -0.452(541) & 82.90(3441) & 68.23(3271) \\
\cmidrule(lr){2-9}
 & \multirow{3}{*}{XGBoost} & Sequential K-Fold & 0.977(1) & 26.26(47) & 18.42(45) & 0.464(210) & 104.23(958) & 73.82(811) \\
 &  & Custom TSCV & 0.964(4) & 32.11(233) & 22.59(150) & 0.460(184) & 109.79(2021) & 76.93(1820) \\
 &  & Blocked TSCV & 0.986(4) & 20.65(280) & 14.40(213) & -0.527(817) & 72.80(1432) & 57.27(1186) \\
\midrule
\multirow{15}{*}{B} & \multirow{3}{*}{Ridge} & Sequential K-Fold & 0.842(7) & 68.74(196) & 50.88(241) & 0.600(162) & 90.79(1120) & 71.03(1227) \\
 &  & Custom TSCV & 0.856(13) & 64.47(391) & 46.93(266) & 0.628(91) & 92.22(1497) & 71.48(1001) \\
 &  & Blocked TSCV & 0.883(33) & 58.26(670) & 40.97(258) & -0.619(1028) & 76.29(1890) & 65.74(2039) \\
\cmidrule(lr){2-9}
 & \multirow{3}{*}{Lasso} & Sequential K-Fold & 0.813(13) & 74.63(216) & 56.62(241) & 0.641(174) & 85.41(1488) & 66.25(1674) \\
 &  & Custom TSCV & 0.808(6) & 74.30(210) & 56.54(151) & 0.677(57) & 86.28(1207) & 67.20(789) \\
 &  & Blocked TSCV & 0.884(33) & 58.18(672) & 40.90(261) & -1.434(2172) & 83.94(1865) & 74.05(1920) \\
\cmidrule(lr){2-9}
 & \multirow{3}{*}{ElasticNet} & Sequential K-Fold & 0.833(8) & 70.52(201) & 53.11(226) & 0.615(183) & 88.35(1348) & 69.62(1545) \\
 &  & Custom TSCV & 0.828(10) & 70.35(299) & 54.00(210) & 0.666(69) & 87.58(1227) & 69.34(877) \\
 &  & Blocked TSCV & 0.878(33) & 59.58(650) & 42.67(317) & -0.451(906) & 74.02(2221) & 63.00(2373) \\
\cmidrule(lr){2-9}
 & \multirow{3}{*}{Random Forest} & Sequential K-Fold & 0.995(0) & 11.73(75) & 7.82(61) & -0.282(1011) & 154.37(3775) & 111.76(3399) \\
 &  & Custom TSCV & 0.969(3) & 30.07(201) & 21.15(128) & 0.498(139) & 105.93(857) & 83.97(923) \\
 &  & Blocked TSCV & 0.991(3) & 16.00(277) & 10.95(213) & -1.267(1832) & 97.75(4605) & 79.15(4011) \\
\cmidrule(lr){2-9}
 & \multirow{3}{*}{XGBoost} & Sequential K-Fold & 0.965(2) & 32.52(106) & 20.88(82) & 0.595(200) & 91.65(2014) & 68.33(1767) \\
 &  & Custom TSCV & 0.960(4) & 33.98(229) & 21.86(98) & 0.625(185) & 92.40(3021) & 65.56(2270) \\
 &  & Blocked TSCV & 0.977(4) & 25.99(233) & 17.28(188) & -0.262(852) & 63.53(1595) & 48.62(1087) \\
\midrule
\multirow{15}{*}{C} & \multirow{3}{*}{Ridge} & Sequential K-Fold & 0.855(8) & 65.82(268) & 49.66(251) & 0.641(149) & 85.87(1342) & 67.29(1192) \\
 &  & Custom TSCV & 0.872(11) & 60.62(347) & 45.59(250) & 0.633(99) & 91.19(1418) & 71.11(1016) \\
 &  & Blocked TSCV & 0.904(10) & 53.26(176) & 39.44(152) & -1.287(1304) & 86.47(966) & 73.62(1255) \\
\cmidrule(lr){2-9}
 & \multirow{3}{*}{Lasso} & Sequential K-Fold & 0.828(12) & 71.61(241) & 55.01(235) & 0.675(153) & 81.27(1552) & 63.63(1525) \\
 &  & Custom TSCV & 0.838(5) & 68.38(178) & 52.68(114) & 0.689(64) & 84.63(1288) & 66.30(980) \\
 &  & Blocked TSCV & 0.905(10) & 53.17(178) & 39.35(156) & -1.241(1249) & 86.02(1022) & 73.26(1282) \\
\cmidrule(lr){2-9}
 & \multirow{3}{*}{ElasticNet} & Sequential K-Fold & 0.848(9) & 67.46(275) & 51.54(241) & 0.662(151) & 82.92(1459) & 65.39(1336) \\
 &  & Custom TSCV & 0.851(8) & 65.55(260) & 50.59(183) & 0.675(69) & 86.39(1290) & 68.16(983) \\
 &  & Blocked TSCV & 0.886(9) & 58.29(109) & 44.02(223) & -0.723(1016) & 78.17(1611) & 64.74(1978) \\
\cmidrule(lr){2-9}
 & \multirow{3}{*}{Random Forest} & Sequential K-Fold & 0.994(1) & 13.90(90) & 9.57(65) & 0.169(635) & 124.58(3168) & 90.95(2526) \\
 &  & Custom TSCV & 0.992(1) & 15.24(65) & 10.65(66) & 0.600(82) & 96.07(1589) & 74.25(1679) \\
 &  & Blocked TSCV & 0.993(2) & 14.75(246) & 10.21(192) & -0.347(539) & 80.97(3610) & 64.09(3149) \\
\cmidrule(lr){2-9}
 & \multirow{3}{*}{XGBoost} & Sequential K-Fold & 0.967(1) & 31.18(100) & 21.22(74) & 0.751(103) & 71.94(1539) & 52.62(1113) \\
 &  & Custom TSCV & 0.974(2) & 27.20(122) & 18.54(62) & 0.673(119) & 85.82(2115) & 61.55(1794) \\
 &  & Blocked TSCV & 0.982(3) & 23.44(222) & 16.12(175) & 0.009(521) & 59.10(1012) & 46.80(910) \\
\bottomrule
\end{tabular}
\end{threeparttable}
\end{table}
\end{landscape}

\begin{landscape}
\begin{table}[htbp]
\centering
\scriptsize
\caption{Final hold-out evaluation across all three feature variants, full metric set, companion to Figure 13. MLR and the physics baseline are reported once per variant; the five tunable models were refit on the full training set using hyperparameters selected under each cross-validation scheme and evaluated on the common chronological hold-out set.}
\label{tab:holdout_full_all_variants}
\begin{threeparttable}
\resizebox{\linewidth}{!}{%
\begin{tabular}{l l l S[table-format=1.3] S[table-format=3.2] S[table-format=3.2] S[table-format=2.2] S[table-format=2.2]
                        S[table-format=-1.3] S[table-format=3.2] S[table-format=3.2] S[table-format=2.2] S[table-format=2.2]}
\toprule
{Variant} & {Model} & {CV Scheme} & {Train $R^2$} & {Train RMSE} & {Train MAE} & {Train Accuracy (\%)} & {Train PA15 (\%)} & {Test $R^2$} & {Test RMSE} & {Test MAE} & {Test Accuracy (\%)} & {Test PA15 (\%)} \\
\midrule
\multirow{17}{*}{A} & \textit{Physics} & \textit{--} & 0.747 & 87.63 & 66.34 & 87.33 & 69.11 & 0.714 & 86.09 & 70.51 & 86.45 & 67.23 \\
 & \textit{MLR} & \textit{--} & 0.785 & 80.81 & 60.33 & 88.21 & 72.40 & 0.837 & 65.02 & 47.66 & 91.58 & 82.49 \\
\cmidrule(lr){2-13}
 & \multirow{3}{*}{Ridge} & Sequential K-Fold & 0.785 & 80.82 & 60.35 & 88.20 & 72.39 & 0.837 & 65.06 & 47.65 & 91.58 & 82.51 \\
 &  & Custom TSCV & 0.785 & 80.82 & 60.35 & 88.20 & 72.39 & 0.837 & 65.06 & 47.65 & 91.58 & 82.51 \\
 &  & Blocked TSCV & 0.785 & 80.82 & 60.35 & 88.20 & 72.39 & 0.837 & 65.06 & 47.65 & 91.58 & 82.51 \\
\cmidrule(lr){2-13}
 & \multirow{3}{*}{Lasso} & Sequential K-Fold & 0.766 & 84.26 & 64.16 & 87.48 & 69.46 & 0.811 & 70.10 & 52.86 & 90.38 & 77.50 \\
 &  & Custom TSCV & 0.761 & 85.24 & 65.15 & 87.32 & 68.88 & 0.801 & 71.80 & 54.59 & 90.12 & 77.07 \\
 &  & Blocked TSCV & 0.784 & 81.00 & 60.60 & 88.17 & 72.36 & 0.827 & 67.05 & 48.74 & 91.39 & 81.85 \\
\cmidrule(lr){2-13}
 & \multirow{3}{*}{ElasticNet} & Sequential K-Fold & 0.778 & 82.09 & 62.29 & 87.89 & 71.10 & 0.823 & 67.77 & 51.33 & 90.66 & 78.32 \\
 &  & Custom TSCV & 0.763 & 84.73 & 65.49 & 87.37 & 68.78 & 0.802 & 71.60 & 56.97 & 89.31 & 73.88 \\
 &  & Blocked TSCV & 0.767 & 84.17 & 65.03 & 87.47 & 69.34 & 0.807 & 70.84 & 56.33 & 89.44 & 74.11 \\
\cmidrule(lr){2-13}
 & \multirow{3}{*}{Random Forest} & Sequential K-Fold & 0.961 & 34.47 & 24.96 & 95.25 & 95.44 & 0.503 & 113.60 & 88.07 & 80.54 & 62.72 \\
 &  & Custom TSCV & 0.997 & 10.02 & 6.63 & 98.86 & 99.99 & 0.369 & 127.92 & 97.40 & 78.09 & 60.54 \\
 &  & Blocked TSCV & 0.991 & 16.86 & 11.37 & 98.03 & 99.89 & 0.496 & 114.30 & 88.93 & 80.57 & 61.37 \\
\cmidrule(lr){2-13}
 & \multirow{3}{*}{XGBoost} & Sequential K-Fold & 0.973 & 28.80 & 20.03 & 96.21 & 97.45 & -0.004 & 161.39 & 109.52 & 74.30 & 65.62 \\
 &  & Custom TSCV & 0.952 & 38.38 & 26.70 & 95.01 & 94.95 & 0.688 & 89.97 & 71.17 & 85.61 & 67.51 \\
 &  & Blocked TSCV & 0.976 & 27.23 & 18.82 & 96.53 & 98.43 & 0.330 & 131.86 & 93.96 & 79.22 & 65.52 \\
\midrule
\multirow{16}{*}{B} & \textit{MLR} & \textit{--} & 0.835 & 70.68 & 52.60 & 89.94 & 77.30 & 0.853 & 61.82 & 40.91 & 92.91 & 89.74 \\
\cmidrule(lr){2-13}
 & \multirow{3}{*}{Ridge} & Sequential K-Fold & 0.835 & 70.69 & 52.62 & 89.94 & 77.30 & 0.852 & 61.94 & 41.04 & 92.88 & 89.64 \\
 &  & Custom TSCV & 0.835 & 70.69 & 52.62 & 89.94 & 77.30 & 0.852 & 61.94 & 41.04 & 92.88 & 89.64 \\
 &  & Blocked TSCV & 0.835 & 70.69 & 52.62 & 89.94 & 77.30 & 0.852 & 61.94 & 41.04 & 92.88 & 89.64 \\
\cmidrule(lr){2-13}
 & \multirow{3}{*}{Lasso} & Sequential K-Fold & 0.809 & 76.05 & 57.73 & 88.94 & 72.34 & 0.840 & 64.49 & 43.46 & 92.63 & 88.59 \\
 &  & Custom TSCV & 0.803 & 77.39 & 59.25 & 88.69 & 71.27 & 0.829 & 66.60 & 46.12 & 92.22 & 87.53 \\
 &  & Blocked TSCV & 0.835 & 70.68 & 52.60 & 89.94 & 77.30 & 0.853 & 61.83 & 40.92 & 92.90 & 89.73 \\
\cmidrule(lr){2-13}
 & \multirow{3}{*}{ElasticNet} & Sequential K-Fold & 0.828 & 72.29 & 54.78 & 89.58 & 75.58 & 0.853 & 61.70 & 40.82 & 93.18 & 90.78 \\
 &  & Custom TSCV & 0.812 & 75.53 & 58.44 & 88.94 & 72.36 & 0.838 & 64.78 & 46.38 & 91.91 & 85.33 \\
 &  & Blocked TSCV & 0.832 & 71.49 & 53.85 & 89.75 & 76.49 & 0.854 & 61.66 & 41.08 & 93.09 & 91.22 \\
\cmidrule(lr){2-13}
 & \multirow{3}{*}{Random Forest} & Sequential K-Fold & 0.996 & 11.72 & 7.83 & 98.64 & 99.95 & 0.315 & 133.30 & 99.50 & 80.02 & 56.11 \\
 &  & Custom TSCV & 0.961 & 34.42 & 24.29 & 95.38 & 95.60 & 0.575 & 104.97 & 83.43 & 81.85 & 63.72 \\
 &  & Blocked TSCV & 0.990 & 17.04 & 11.51 & 98.00 & 99.88 & 0.385 & 126.33 & 98.44 & 78.24 & 58.24 \\
\cmidrule(lr){2-13}
 & \multirow{3}{*}{XGBoost} & Sequential K-Fold & 0.960 & 34.71 & 21.98 & 96.01 & 97.39 & 0.752 & 80.30 & 62.38 & 87.31 & 72.13 \\
 &  & Custom TSCV & 0.954 & 37.24 & 23.60 & 95.68 & 96.77 & 0.799 & 72.21 & 55.85 & 88.75 & 76.48 \\
 &  & Blocked TSCV & 0.960 & 35.02 & 22.26 & 95.94 & 97.26 & 0.771 & 77.17 & 59.93 & 87.82 & 73.30 \\
\midrule
\multirow{16}{*}{C} & \textit{MLR} & \textit{--} & 0.849 & 67.67 & 51.12 & 90.21 & 77.84 & 0.870 & 57.98 & 40.99 & 92.85 & 89.51 \\
\cmidrule(lr){2-13}
 & \multirow{3}{*}{Ridge} & Sequential K-Fold & 0.849 & 67.69 & 51.15 & 90.21 & 77.83 & 0.870 & 58.05 & 41.09 & 92.83 & 89.42 \\
 &  & Custom TSCV & 0.849 & 67.69 & 51.15 & 90.21 & 77.83 & 0.870 & 58.05 & 41.09 & 92.83 & 89.42 \\
 &  & Blocked TSCV & 0.849 & 67.69 & 51.15 & 90.21 & 77.83 & 0.870 & 58.05 & 41.09 & 92.83 & 89.42 \\
\cmidrule(lr){2-13}
 & \multirow{3}{*}{Lasso} & Sequential K-Fold & 0.825 & 72.92 & 55.95 & 89.28 & 73.34 & 0.858 & 60.70 & 43.89 & 92.53 & 88.24 \\
 &  & Custom TSCV & 0.825 & 72.92 & 55.95 & 89.28 & 73.34 & 0.858 & 60.70 & 43.89 & 92.53 & 88.24 \\
 &  & Blocked TSCV & 0.849 & 67.68 & 51.13 & 90.21 & 77.84 & 0.870 & 58.00 & 41.02 & 92.84 & 89.48 \\
\cmidrule(lr){2-13}
 & \multirow{3}{*}{ElasticNet} & Sequential K-Fold & 0.843 & 69.11 & 52.91 & 89.87 & 76.37 & 0.869 & 58.34 & 41.42 & 93.02 & 89.18 \\
 &  & Custom TSCV & 0.832 & 71.38 & 55.21 & 89.46 & 74.38 & 0.861 & 60.10 & 44.47 & 92.45 & 87.64 \\
 &  & Blocked TSCV & 0.836 & 70.53 & 54.49 & 89.61 & 75.19 & 0.866 & 59.01 & 43.28 & 92.70 & 88.45 \\
\cmidrule(lr){2-13}
 & \multirow{3}{*}{Random Forest} & Sequential K-Fold & 0.994 & 14.01 & 9.69 & 98.26 & 99.89 & 0.567 & 106.03 & 82.78 & 83.23 & 61.76 \\
 &  & Custom TSCV & 0.992 & 15.31 & 10.49 & 98.17 & 99.92 & 0.485 & 115.63 & 88.93 & 80.15 & 62.61 \\
 &  & Blocked TSCV & 0.992 & 15.59 & 10.66 & 98.14 & 99.91 & 0.458 & 118.58 & 90.76 & 79.74 & 62.37 \\
\cmidrule(lr){2-13}
 & \multirow{3}{*}{XGBoost} & Sequential K-Fold & 0.965 & 32.79 & 22.31 & 95.92 & 97.37 & 0.852 & 62.08 & 50.78 & 90.15 & 78.38 \\
 &  & Custom TSCV & 0.970 & 30.27 & 20.44 & 96.27 & 98.07 & 0.789 & 74.05 & 58.35 & 88.43 & 72.39 \\
 &  & Blocked TSCV & 0.968 & 31.04 & 21.20 & 96.12 & 97.83 & 0.831 & 66.17 & 54.19 & 89.33 & 76.32 \\
\bottomrule
\end{tabular}
}
\begin{tablenotes}
\small
\item Units: RMSE and MAE in L h$^{-1}$; Accuracy and PA15 in percent.
\end{tablenotes}
\end{threeparttable}
\end{table}
\end{landscape}

\end{document}